\documentclass[10pt,journal,compsoc]{IEEEtran}

\usepackage{amsmath}
\usepackage{amssymb}
\usepackage{amsthm}
\usepackage{booktabs}
\usepackage{enumitem}
\usepackage{graphicx}
\usepackage{makecell}
\usepackage{mathtools}
\usepackage{marvosym}
\usepackage{microtype}
\usepackage{multicol}
\usepackage{multirow}
\usepackage{ragged2e}
\usepackage{subcaption}
\usepackage{tabularx}
\usepackage[table,dvipsnames]{xcolor}
\usepackage{tikz}
\usepackage{wrapfig}
\usepackage{pifont}
\usepackage{xspace}
\usepackage{algorithm}
\usepackage{algorithmic}
\usepackage[normalem]{ulem}

\usepackage[colorlinks,urlcolor=magenta]{hyperref}

\usepackage[capitalize]{cleveref}
\crefname{section}{Sec.}{Secs.}
\Crefname{section}{Section}{Sections}
\Crefname{table}{Table}{Tables}
\crefname{table}{Tab.}{Tabs.}

\definecolor{sf_blue}{RGB}{66,133,244}
\definecolor{sf_red}{RGB}{231,66,52}
\definecolor{sf_yellow}{RGB}{251,189,5}
\definecolor{sf_green}{RGB}{51,168,82}
\definecolor{sf_gray}{RGB}{165,165,165}

\newcommand{\cmark}{\ding{51}}
\newcommand{\xmark}{\ding{55}}

\ifCLASSOPTIONcompsoc
  \usepackage[nocompress]{cite}
\else
  \usepackage{cite}
\fi

\begin{document}

\title{Enhanced Spatiotemporal Consistency for Image-to-LiDAR Data Pretraining}

\author{Xiang~Xu$^*$, Lingdong~Kong$^*$, Hui~Shuai, Wenwei~Zhang, Liang~Pan, Kai~Chen, Ziwei~Liu, Qingshan~Liu
\IEEEcompsocitemizethanks{
\IEEEcompsocthanksitem X. Xu is with the College of Computer Science and Technology, Nanjing University of Aeronautics and Astronautics, Nanjing, China.
\IEEEcompsocthanksitem L. Kong is with the School of Computing, Department of Computer Science, National University of Singapore, and CNRS@CREATE, Singapore.
\IEEEcompsocthanksitem H. Shuai and Q. Liu are with the School of Computer Science, Nanjing University of Posts and Telecommunications, Nanjing, China.
\IEEEcompsocthanksitem W. Zhang, L. Pan, and K. Chen are with Shanghai AI Laboratory, Shanghai, China.
\IEEEcompsocthanksitem Z. Liu is with S-Lab, Nanyang Technological University, Singapore.
\IEEEcompsocthanksitem The corresponding author is Q. Liu (E-mail: \url{qsliu@njupt.edu.cn}).
\IEEEcompsocthanksitem X. Xu and L. Kong contributed equally to this work.}}

\IEEEtitleabstractindextext{
    \begin{abstract}
    LiDAR representation learning has emerged as a promising approach to reducing reliance on costly and labor-intensive human annotations. While existing methods primarily focus on spatial alignment between LiDAR and camera sensors, they often overlook the temporal dynamics critical for capturing motion and scene continuity in driving scenarios. To address this limitation, we propose \textbf{SuperFlow++}, a novel framework that integrates spatiotemporal cues in both pretraining and downstream tasks using consecutive LiDAR-camera pairs. SuperFlow++ introduces four key components: \textbf{(1)} a view consistency alignment module to unify semantic information across camera views, \textbf{(2)} a dense-to-sparse consistency regularization mechanism to enhance feature robustness across varying point cloud densities, \textbf{(3)} a flow-based contrastive learning approach that models temporal relationships for improved scene understanding, and \textbf{(4)} a temporal voting strategy that propagates semantic information across LiDAR scans to improve prediction consistency. Extensive evaluations on 11 heterogeneous LiDAR datasets demonstrate that SuperFlow++ outperforms state-of-the-art methods across diverse tasks and driving conditions. Furthermore, by scaling both 2D and 3D backbones during pretraining, we uncover emergent properties that provide deeper insights into developing scalable 3D foundation models. With strong generalizability and computational efficiency, SuperFlow++ establishes a new benchmark for data-efficient LiDAR-based perception in autonomous driving. The code is publicly available at \url{https://github.com/Xiangxu-0103/SuperFlow}.
\end{abstract}


}

\maketitle

\IEEEdisplaynontitleabstractindextext
\IEEEpeerreviewmaketitle

\section{Introduction}
\label{sec:introduction}

Robust perception is essential for autonomous driving, forming the basis for safe and efficient navigation. Advances in sensing technologies, such as LiDAR (Light Detection and Ranging) and surround-view cameras, have greatly improved 3D scene understanding, enabling comprehensive and precise environmental perception \cite{badue2021survey, rizzoli2022survey, sun2024lidarseg}.

Despite these advances, achieving high-performing 3D perception remains challenging due to the heavy reliance on extensive labeled data and high computational costs \cite{gao2021survey, liu2024survey}. Compared to 2D images, annotating 3D point clouds is more expensive and labor-intensive, posing a major scalability bottleneck for existing methods \cite{muhammad2020survey, xiao2023survey, zhang2024survey, geiger2012kitti}. To mitigate this challenge, data representation learning has emerged as a promising alternative \cite{bengio2013survey}. By designing effective pretraining objectives, models can extract meaningful features from raw, unlabeled data, reducing dependence on large-scale annotations while improving downstream performance \cite{lekhac2020survey}.

Following this paradigm, Sautier \textit{et al.} \cite{sautier2022slidr} introduced SLidR, a framework that transfers knowledge from 2D pretrained backbones, such as MoCo \cite{chen2020moCoV2} and DINO \cite{oquab2023dinov2}, to LiDAR point clouds. SLidR leverages a superpixel-driven contrastive learning mechanism to align camera and LiDAR features. Subsequent works have refined this approach with techniques such as class balancing \cite{mahmoud2023st-slidr}, hybrid-view distillation \cite{zhang2024hvdistill,xu2025limoe}, semantic superpixels \cite{liu2023seal,kong2025largead}, and representation distillation \cite{puy2024scalar}. While these methods have substantially advanced LiDAR representation learning, they primarily focus on spatial alignment. However, they largely overlook the temporal dynamics inherent in LiDAR data \cite{caesar2020nuScenes, behley2019SemanticKITTI}, which are crucial for achieving robust, context-aware 3D perception in autonomous driving \cite{nunes2023tarl,wu2023stssl}.

Recently, our preliminary work, SuperFlow \cite{xu2024superflow}, introduced a semantic-flow framework for consecutive LiDAR scans, enforcing dynamic consistency and improving LiDAR representation learning. It comprises three key components: (1) \textbf{View Consistency Alignment}, which refines the segmentation head of vision foundation models (VFMs) \cite{zhang2023openSeeD,zou2023seem,kirillov2023sam} using predefined prompts, generating semantic-aware superpixels across camera views to align semantic flow across timestamps; (2) \textbf{Dense-to-Sparse Consistency Regularization}, which enforces feature consistency between dense and sparse point clouds, mitigating the impact of irregular LiDAR point distributions; and (3) \textbf{Flow-Based Contrastive Learning}, which incorporates spatial and temporal contrastive learning to jointly capture static structures and dynamic motion patterns for robust autonomous perception.

However, several open challenges remain. First, the complexity and variability of real-world driving environments introduce factors such as sensor noise, motion blur, and adverse weather conditions, all of which can substantially degrade feature quality and hinder downstream performance \cite{kong2023robo3D,xiao2023semanticSTF}. Although SuperFlow \cite{xu2024superflow} propagates LiDAR features across frames to enforce temporal consistency, its learned representations remain vulnerable to these disturbances. Second, most downstream pipelines continue to operate in a per-frame manner, ignoring the inherent temporal continuity of LiDAR sequences. This lack of temporal exploitation often results in inconsistent predictions across frames, ultimately limiting the robustness and generalizability of SuperFlow in challenging autonomous driving scenarios.

To address these limitations, we propose \textbf{SuperFlow++}, an enhanced framework that reinforces temporal modeling from pretraining to downstream tasks. Building upon our preliminary work \cite{xu2024superflow}, it introduces the following extensions:
\begin{itemize}
    \item \textbf{Refined Language-Guided Semantic Superpixels.} While our previous work utilizes VFMs to generate semantic superpixels, SuperFlow++ enhances this process by incorporating LiDAR-camera calibration. This refinement ensures better alignment of superpixels across overlapping camera views, reducing semantic inconsistencies caused by viewpoint variations. By unifying semantic cues across multiple cameras, SuperFlow++ provides a more stable and structured representation of scene elements, improving cross-view consistency and facilitating more accurate LiDAR-camera fusion.

    \item \textbf{Cross-Sensor Temporal Contrastive Learning.} To better capture the evolution of dynamic scenes, we introduce a cross-sensor temporal contrastive learning module that explicitly aligns temporal cues between cameras and LiDAR across different timestamps. Unlike prior approaches that focus solely on intra-modal temporal learning, our method enforces feature consistency across both spatial and temporal domains, ensuring that motion patterns and semantic structures remain coherent over time. This enables the model to leverage complementary information from different sensors, improving its robustness to motion-induced distortions and sensor noise in complex driving environments.

    \item \textbf{Temporal Voting for Downstream Tasks.} To address the challenge of per-frame inconsistencies in downstream tasks, SuperFlow++ introduces a temporal voting strategy that aggregates semantic predictions from consecutive LiDAR frames. By integrating contextual from neighboring frames into current frame predictions, this strategy mitigates transient misclassifications, enhances robustness to occlusions, and smooths out segmentation artifacts caused by sparsity variations. This mechanism not only refines per-frame segmentation accuracy but also ensures temporal stability, which is critical for safety-critical applications such as autonomous driving.
\end{itemize}
By integrating these newly proposed components into SuperFlow \cite{xu2024superflow}, SuperFlow++ fosters a semantically and temporally enriched feature space, significantly improving data representation learning. Our framework demonstrates substantial performance gains over the baseline \cite{xu2024superflow} across a wide range of downstream tasks, including segmentation, detection, and robustness evaluation. Furthermore, we investigate the scalability of both 2D and 3D network backbones during pretraining and evaluate their impact on semi-supervised learning, providing insights into the development of unified and scalable 3D perception models for diverse real-world applications.

In summary, our key contributions are as follows:
\begin{itemize}
    \item We present \textbf{SuperFlow++}, an enhanced framework that enforces temporal consistency across consecutive LiDAR scans, integrating it seamlessly into both pretraining and downstream tasks.

    \item For pretraining, we introduce three key components: a view consistency alignment module, a dense-to-sparse consistency regularization mechanism, and a flow-based contrastive learning module. These components collectively unify the semantic feature space and enhance feature representation learning by effectively capturing temporal dynamics.

    \item For downstream tasks, we introduce a temporal voting strategy that aggregates semantic information across consecutive frames, ensuring consistency and reducing transient misclassifications in LiDAR-based perception.

    \item Extensive experiments on multiple LiDAR benchmarks validate the effectiveness of SuperFlow++. Moreover, we explore the scalability of both 2D and 3D backbones and analyze their impact on semi-supervised learning, providing insights for developing more robust and efficient LiDAR representation learning frameworks in real-world applications.
\end{itemize}

The remainder of this paper is structured as follows. \cref{sec:related_work} reviews related literature on autonomous driving perception and pretraining, as well as temporal modeling strategies. \cref{sec:revisit} provides an overview of image-to-LiDAR contrastive learning. \cref{sec:methodology} details the proposed spatiotemporal contrastive learning framework. \cref{sec:experiments} presents experimental evaluations and results. Finally, \cref{sec:conclusion} concludes the paper with discussions on future research directions.

\begin{table}
    \centering
    \caption{Categorization of LiDAR representation learning.}
    \vspace{-0.2cm}
    \label{tab:category}
    \begin{tabular}{c|c|c}
        \toprule
        \textbf{Input Modality} & \textbf{Objective} & \textbf{References}
        \\\midrule\midrule
        \multirow{4}{*}{LiDAR Sensor} & Contrastive Learning & \cite{xie2020pointcontrast,zhang2021depthcontrast}
        \\
        & Masked Modeling & \cite{hess2023masked,krispel2024maeli}
        \\
        & Reconstruction & \cite{michele2024saluda,boulch2023also}
        \\
        & Spatiotemporal & \cite{huang2021strl,nunes2023tarl,wu2023stssl,sautier2024bevcontrast}
        \\\midrule
        \multirow{3}{*}{Image \& LiDAR} & Contrastive Learning & \cite{sautier2022slidr,mahmoud2023st-slidr,zhang2024hvdistill,chen2024csc}
        \\
        & Distillation & \cite{puy2024scalar}
        \\
        & Spatiotemporal & \cite{liu2023seal,xu2024superflow}
        \\\bottomrule
    \end{tabular}
\end{table}

\section{Related Work}
\label{sec:related_work}

In this section, we provide a comprehensive review of the literature, reorganizing the discussion into four interconnected themes: LiDAR-based 3D perception, data-efficient 3D learning, cross-sensor representation learning, and spatiotemporal (4D) learning. This reformulation reflects the progression from traditional methods to advanced techniques leveraging temporal and cross-modal consistency.

\subsection{3D Perception with LiDAR Sensors}

LiDAR sensors have become fundamental for 3D perception systems, offering precise geometric scene understanding \cite{triess2021survey, behley2019SemanticKITTI}. However, the sparsity and unordered nature of point clouds pose challenges, necessitating effective representations for downstream tasks \cite{uecker2022analyzing, hu2021sensatUrban}. Common representations include raw point-based processing \cite{qi2017pointnet, qi2017pointnet++, hu2020randlanet, thomas2019kpconv, shuai2021baflac}, sparse voxel grids \cite{choy2019minkowski, zhu2021cylindrical, tang2020searching, hong2021dsnet}, bird’s-eye view (BEV) projections \cite{zhou2020polarNet, chen2021polarStream, zhou2021panoptic}, and range-view encodings \cite{milioto2019rangenet++, xu2020squeezesegv3, cortinhal2020salsanext, xu2025frnet, kong2023rangeformer}. Multi-view fusion techniques \cite{xu2021rpvnet, cheng2021af2S3Net, xu2023multiview, liong2020amvNet, liu2023uniseg} further enhance scene understanding by integrating complementary perspectives. Despite their success, these approaches are highly reliant on labeled data, posing scalability concerns \cite{gao2021survey}. Consequently, research has shifted toward data-efficient learning paradigms such as self-supervised, weakly supervised, and semi-supervised learning to mitigate annotation burdens. Our work follows this direction, proposing a multi-modal pretraining framework that enhances 3D perception without requiring extensive manual labels.

\subsection{Data-Efficient 3D Learning}

Given the high annotation cost of LiDAR data, research has explored weakly supervised and semi-supervised learning strategies \cite{gao2021survey, kong2023conDA}. Weak supervision techniques, including sparse seed annotations \cite{shi2022weak, hu2022sqn}, scribble-based labels \cite{unal2022scribbleKITTI}, and active learning \cite{liu2022less, xie2023annotator}, have shown promise in reducing labeling costs. Semi-supervised methods, such as consistency-based learning \cite{tarvainen2017MeanTeacher} and data augmentation-based strategies \cite{ kong2023lasermix, kong2025lasermix++}, leverage vast amounts of unlabeled LiDAR data to improve model generalization. While these methods achieve strong performance, they often lack alignment with existing pretraining paradigms, limiting their scalability to large-scale applications. To bridge this gap, our work investigates representation learning with data-efficient strategies, ensuring compatibility between pretraining and semi-supervised learning. This enhances scalability and adaptability while maintaining robust performance across diverse driving environments.

\subsection{Cross-Sensor Data Pretraining}

Early 3D pretraining approaches borrowed heavily from 2D paradigms, including contrastive learning \cite{xie2020pointcontrast, zhang2021depthcontrast}, masked modeling \cite{hess2023masked, krispel2024maeli}, and reconstruction-based learning \cite{michele2024saluda, boulch2023also}. However, these methods primarily focused on single-modal point clouds, missing the complementary advantages of cross-modal sensor fusion. To address this limitation, recent works have explored LiDAR-camera joint pretraining. SLidR \cite{sautier2022slidr} introduced superpixel-driven contrastive objectives to align LiDAR and camera features. ST-SLidR \cite{mahmoud2023st-slidr} and HVDistill \cite{zhang2024hvdistill} further refined this approach by enhancing feature alignment and optimizing distillation pipelines. Seal \cite{liu2023seal} integrated vision foundation models \cite{kirillov2023sam, zou2023seem} to improve robustness, while ScaLR \cite{puy2024scalar} focused on pretraining efficiency. Despite these advancements, temporal dependencies in LiDAR-camera sequences remain underexplored, limiting motion understanding. Our work addresses this gap by enforcing temporal alignment and cross-modal consistency, capturing richer spatiotemporal representations for more robust scene understanding. An overview of existing LiDAR representation learning methods is presented in \cref{tab:category}.

\subsection{Spatiotemporal (4D) Representation Learning}

Exploiting temporal information from LiDAR scans has emerged as a promising strategy for capturing scene dynamics and motion \cite{aygun2021pls4d, hong20224dDSNet, shi2020spsequencenet, duerr2020lidar}. While 4D representation learning has been widely studied in small-scale scenarios (\textit{e.g.}, human or object tracking) \cite{sheng2023point, liu2023leaf, shen2023masked, zhang2023complete, chen20224dcontrast}, its adaptation to large-scale driving scenes remains underexplored. Early works like STRL \cite{huang2021strl} introduced spatiotemporal invariance through sequence augmentations, while TARL \cite{nunes2023tarl} and STSSL \cite{wu2023stssl} employed clustering-based frame alignment using techniques like RANSAC \cite{foschler1981ransac}, Patchwork \cite{lim2021patchwork}, and HDBSCAN \cite{ester1996dbscan}. More recently, BEVContrast \cite{sautier2024bevcontrast} extended contrastive learning to BEV maps, providing a more practical solution for autonomous driving. However, many existing 4D learning methods rely on heuristic clustering, which lacks adaptability to real-world density variations. To overcome this, our approach integrates LiDAR-camera correspondences with superpixel-based pretraining objectives, capturing both spatial and temporal dynamics. This ensures a scalable and generalizable 4D perception framework, advancing temporal feature learning in large-scale autonomous driving scenarios.

\section{Image-to-LiDAR Data Pretraining}
\label{sec:revisit}

In this section, we introduce the standard framework for image-to-LiDAR pretraining, aimed at transferring prior knowledge from images to point clouds. We begin with an overview of the calibration between camera and LiDAR sensors (\cref{sec:calibration}), followed by the generation of superpixels and superpoints (\cref{sec:sp_generation}), and conclude with the introduction of the cross-sensor contrastive objective (\cref{sec:contrastive}).

\subsection{Camera-to-LiDAR Calibration}
\label{sec:calibration}

Modern automotive systems are increasingly equipped with a combination of one LiDAR sensor and multiple cameras to achieve precise perception and robust interpretation of the surrounding environment. This multimodal configuration leverages the complementary strengths of LiDAR and cameras: LiDAR provides accurate depth and geometric information, while cameras capture rich texture and color details. To enable seamless integration and effective use of these data sources, precise calibration between LiDAR and camera sensors is a critical prerequisite.

Calibration ensures the spatial alignment between LiDAR and camera data by establishing a consistent mapping from the 3D LiDAR point cloud to the 2D image planes. Specifically, consider a point cloud $\mathcal{P} = \{\mathbf{p}_i, \mathbf{e}_i | i = 1, ..., N\}$ comprising $N$ points captured by the LiDAR sensor. Here, $\mathbf{p}_i = (x_i, y_i, z_i) \in \mathbb{R}^3$ denotes the 3D coordinate of the $i$-th point, and $\mathbf{e}_i \in \mathbb{R}^L$ represents its associated feature (\textit{e.g.}, intensity, elongation). Concurrently, a set of $V$ images $\{\mathcal{I}_1, ..., \mathcal{I}_V\}$ is captured by $V$ cameras, where each image $\mathcal{I}_j \in \mathbb{R}^{H \times W \times 3}$ has height $H$ and width $W$. To project a 3D LiDAR point $\mathbf{p}_i$ onto the $j$-th camera's image plane at pixel coordinates $(u_i^j, v_i^j)$, the following transformation is applied:
\begin{equation}
    \label{eq:projection}
    [u_i^j, v_i^j, 1]^\text{T} = \frac{1}{z_i} \times \Gamma_K^j \times \Gamma_{l \to c}^j \times [x_i, y_i, z_i]^\text{T}~,
\end{equation}
where $\Gamma_K^j$ is the intrinsic matrix of the $j$-th camera, and $\Gamma_{l \to c}^j$ is the extrinsic transformation matrix that defines the rigid-body transformation between the LiDAR coordinate system and the $j$-th camera's coordinate system.

\subsection{Superpixel \& Superpoint Generation}
\label{sec:sp_generation}

To establish meaningful correlations between LiDAR points and camera images, previous works \cite{sautier2022slidr,mahmoud2023st-slidr,liu2023seal} leverage superpixels -- contiguous regions in an image that group pixels with similar properties, such as color, intensity or texture. These superpixels serve as compact representations, effectively bridging the dense, high-resolution information in 2D images with the sparse 3D LiDAR data.

SLidR \cite{sautier2022slidr} employed the unsupervised SLIC algorithm \cite{achanta2012slic} to generate superpixels. SLIC \cite{achanta2012slic} clusters pixels in a five-dimensional space combining spatial coordinates and color values, resulting in regions that adhere closely to object boundaries while ensuring computational efficiency. However, it tends to over-segment images, creating a larger number of smaller regions that may lack semantic coherence.

Recent advancements in Vision Foundation Models (VFMs) \cite{kirillov2023sam,zhang2023openSeeD,zou2023seem,zou2023xdecoder} have paved the way for generating instance superpixels. Unlike traditional methods, VFMs leverage deep neural networks pretrained on extensive datasets to capture high-level semantic features. These methods produce superpixels that not only align with object boundaries but also encapsulate meaningful semantic information. Such superpixels, for instance, correspond to entire objects for functionally relevant regions rather than focusing solely on low-level visual attributes.

For a given image, the superpixel set is denoted as $\mathcal{O}_c = \{\mathbf{o}_c^1, ..., \mathbf{o}_c^M\}$, where $M$ represents the number of superpixel regions. Using the projection defined in \cref{eq:projection}, each LiDAR point in the 3D point cloud can be mapped to the corresponding location on the image plane. This allows LiDAR points to be associated with specific superpixel regions, thereby forming the corresponding superpoint set $\mathcal{O}_p = \{\mathbf{o}_p^1, ..., \mathbf{o}_p^M\}$. Each superpoint represents a group of LiDAR points spatially aligned with a superpixel.

\begin{figure*}[t]
    \centering
    \begin{subfigure}[b]{0.306\textwidth}
        \centering
        \includegraphics[width=\textwidth]{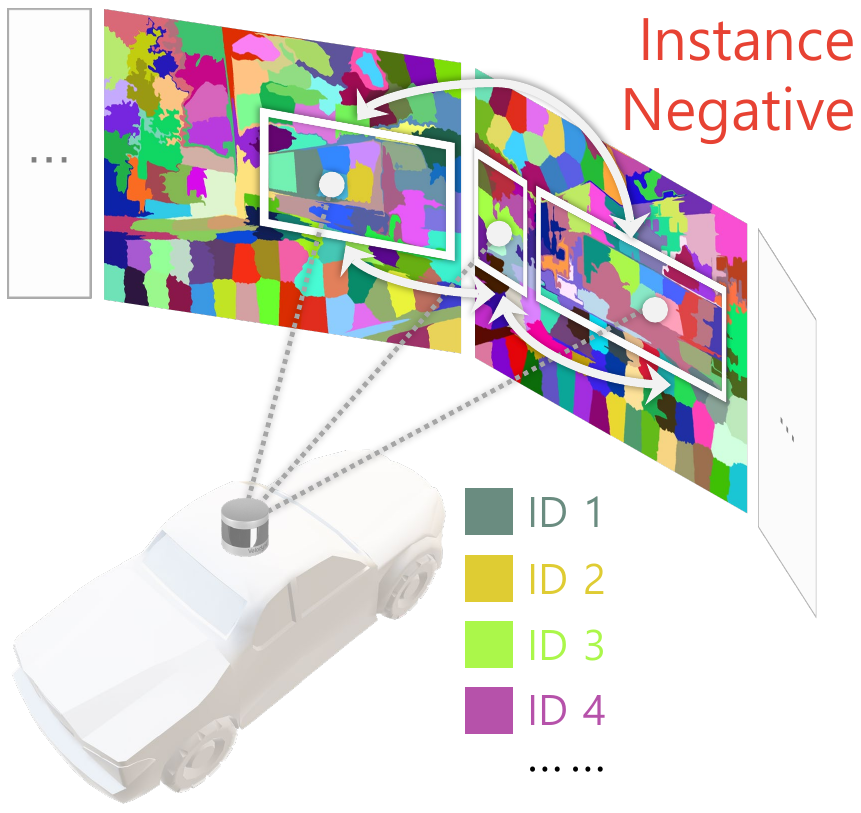}
        \caption{Heuristic}
        \label{fig:slic}
    \end{subfigure}
    ~~
    \begin{subfigure}[b]{0.306\textwidth}
        \centering
        \includegraphics[width=\textwidth]{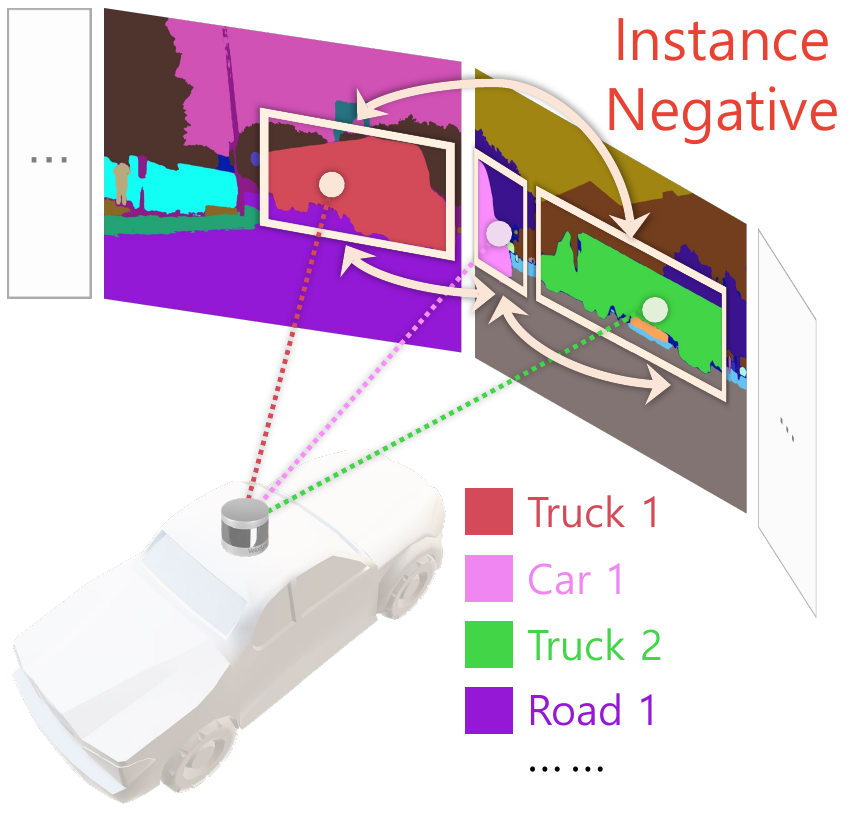}
        \caption{Class Agnostic}
        \label{fig:seem}
    \end{subfigure}
    ~~
    \begin{subfigure}[b]{0.306\textwidth}
        \centering
        \includegraphics[width=\textwidth]{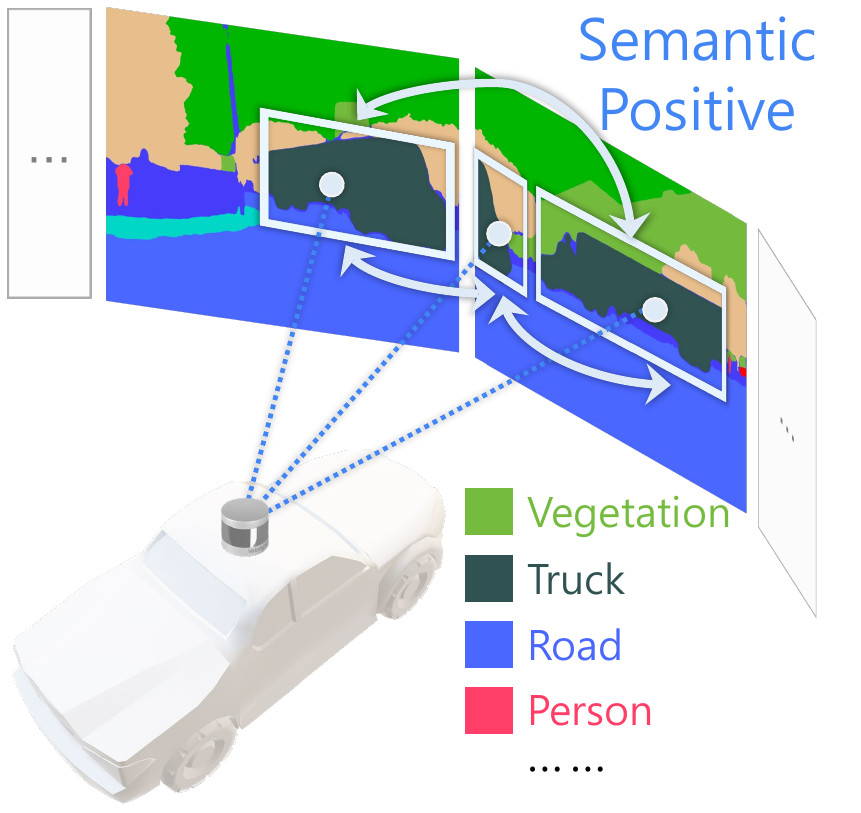}
        \caption{View Consistent}
        \label{fig:vsa}
    \end{subfigure}
    \caption{\textbf{Comparisons of different superpixels.} (a) Class-agnostic superpixels generated by the unsupervised SLIC \cite{achanta2012slic} algorithm. (b) Class-agnostic semantic superpixels generated by vision foundation models (VFMs) \cite{zou2023seem,zhang2023openSeeD,zou2023xdecoder}. (c) View-consistent semantic superpixels generated by our view consistency alignment module.}
    \label{fig:superpixels}
\end{figure*}

\subsection{Cross-Sensor Contrastive Learning}
\label{sec:contrastive}

Let $\mathcal{F}_{\theta_p}: \mathbb{R}^{N \times (3+L)} \to \mathbb{R}^{N \times D}$ denote a 3D backbone network with trainable parameters $\theta_p$, which processes the LiDAR point cloud $\mathcal{P}$ and outputs $D$-dimensional feature representations. Similarly, let $\mathcal{G}_{\theta_i}: \mathbb{R}^{H \times W \times 3} \to \mathbb{R}^{\frac{H}{S} \times \frac{W}{S} \times E}$ represent an image backbone network with pretrained parameters $\theta_i$. This network processes an image $\mathcal{I}$ to produce an $E$-dimensional feature map with spatial dimensions $\frac{H}{S} \times \frac{W}{S}$, where $S$ is the stride. To transfer knowledge from the pretrained image backbone $\mathcal{G}_{\theta_i}$ to the 3D backbone $\mathcal{F}_{\theta_p}$ and enable cross-modal feature alignment. projection heads $\mathcal{H}_{\omega_p}: \mathbb{R}^{N \times D} \to \mathbb{R}^{N \times C}$ and $\mathcal{H}_{\omega_i}: \mathbb{R}^{\frac{H}{S} \times \frac{W}{S} \times E} \to \mathbb{R}^{H \times W \times C}$ are applied to project point and image features, respectively, into a shared $C$-dimensional space. The projected features are subsequently normalized using $\ell_2$-normalization, ensuring that both modalities are represented on the unit hypersphere in $C$-dimensional space.

To further align features across modalities, superpixels $\mathcal{O}_c$ and superpoints $\mathcal{O}_p$ are used to group the image and point features, respectively, generating the corresponding superpixel embeddings $\mathbf{K} \in \mathbb{R}^{M \times C}$ and superpoint embeddings $\mathbf{Q} \in \mathbb{R}^{M \times C}$. A contrastive loss is defined as follows:
\begin{equation}
    \label{eq:contrastive_loss}
    \mathcal{L}_\text{con}(\mathbf{Q}, \mathbf{K}) = - \frac{1}{M} \sum_{i=1}^M \log \left[ \frac{e^{(\langle \mathbf{q}_i, \mathbf{k}_i \rangle / \tau)}}{\sum_{j=1}^M e^{(\langle \mathbf{q}_i, \mathbf{k}_j \rangle / \tau)}} \right]~,
\end{equation}
where $\langle \cdot, \cdot \rangle$ denotes the dot product measuring similarity between embeddings, and $\tau > 0$ is a temperature parameter that controls the smoothness of the distribution. This loss encourages the embeddings of corresponding superpixels and superpoints to pull closer in the shared feature space while pushing apart embeddings of non-corresponding pairs.

Although this strategy achieved promising performance for cross-sensor feature alignment, it overlooks the temporal cues present across consecutive LiDAR scenes. To address this limitation, we propose an advanced framework that leverages consecutive LiDAR-camera pairs to establish spatiotemporal pretraining objectives, enabling the model to capture both spatial and temporal dynamics for enhanced feature representation.

\section{SuperFlow++}
\label{sec:methodology}

In this section, we provide a detailed overview of the proposed framework. We begin with the plug-and-play view consistency alignment module (\cref{sec:vca}), which enhances spatial coherence across multi-view representations. Next, we introduce dense-to-sparse consistency regularization (\cref{sec:dense_to_sparse}) and flow-based contrastive learning (\cref{sec:spatial_temporal}), which facilitate the modeling of temporal dependencies across frames. Finally, we propose a temporal voting strategy for downstream semantic segmentation, ensuring robust semantic consistency across sequential scenes (\cref{sec:temporal_voting}).

\subsection{View Consistency Alignment}
\label{sec:vca}

\noindent\textbf{Motivation.}
The class-agnostic superpixels $\mathcal{O}_c$ utilized in previous studies \cite{sautier2022slidr,mahmoud2023st-slidr,liu2023seal} are primarily instance-focused and fail to incorporate semantic category information. As highlighted in ST-SLidR \cite{mahmoud2023st-slidr}, these instance-level superpixels often yield inconsistent semantic priors during image-to-LiDAR knowledge transferring, where regions belonging to the same semantic category are fragmented into separate instances. This fragmentation disrupts semantic coherence across modalities, thereby weakening the pretraining supervision and hindering the model’s ability to learn robust and semantically meaningful cross-modal representations.

\noindent\textbf{Superpixel Comparisons.}
\cref{fig:superpixels} compares superpixels generated using the unsupervised SLIC algorithm \cite{achanta2012slic} with those produced by VFMs. SLIC \cite{achanta2012slic} tends to over-segment objects, leading to semantic conflicts as it fails to capture coherent object-level representations. On the other hand, VFMs employ a panoptic segmentation head to generate superpixels that align more closely with object boundaries. However, VFMs still encounter ``self-conflict'' issues in several scenarios (illustrated in \cref{fig:seem}): \ding{172} when the same object spans multiple camera views, portions of the object may be incorrectly classified as negative samples; \ding{173} multiple objects of the same category within a single camera view are mistakenly treated as negative samples; and \ding{174} objects of the same category appearing across different camera views are also categorized as negative samples, despite sharing identical labels. These issues result in inconsistent segmentation both within and across camera views, thereby compromising semantic coherence and transfer quality.

\noindent\textbf{Semantic-Related Superpixel Generation.}
To address the aforementioned issues, we propose a two-stage approach for generating semantic-related superpixels, ensuring semantic consistency across camera views. \textbf{First}, leveraging the generalization capabilities of Contrastive Vision-Language Image Pretraining (CLIP) \cite{radford2021CLIP} in few-shot learning, we enhance existing VFMs \cite{kirillov2023sam,zou2023xdecoder,zou2023seem}. Specifically, we employ CLIP's text encoder and fine-tune the final layer of the VFMs' segmentation head with predefined text prompts. This enables the segmentation head to generate language-guided semantic categories for each pixel, which are then used as superpixels. \textbf{Second}, to address inconsistencies when parts of an object appear across neighboring camera views (especially near image edges), we propose a LiDAR-to-image calibration strategy to refine the superpixel generated in the first stage. Since overlapping regions are shared between neighboring cameras, points within these regions can be projected onto multiple camera views. For these points, we analyze their corresponding superpixel information, including instance areas and semantic categories. When inconsistencies arise across views, we unify their semantics according to the dominant (largest-area) instance. As illustrated in \cref{fig:vsa}, this cross-view unification produces spatially and semantically consistent superpixels, providing more reliable semantic priors for subsequent image-to-LiDAR contrastive learning.

\subsection{D2S: Dense-to-Sparse Consistency Regularization}
\label{sec:dense_to_sparse}

\noindent\textbf{Motivation.}
LiDAR point clouds inherently suffer from sparsity and incompleteness in both foreground objects and background regions, which pose substantial challenges for cross-sensor feature representation learning \cite{kong2023robo3D,unal2022scribbleKITTI,bengio2013survey}. Models capable of learning invariant features with respect to varying point densities are therefore crucial for capturing reliable structural and semantic information in 3D space. Therefore, we integrate multi-sweep LiDAR scans captured within a defined time window to create a denser and more comprehensive point cloud. This denser point cloud acts as a reference, guiding the representation learning by promoting consistency with the original sparse point cloud. This approach not only mitigates the inherent sparsity of LiDAR data but also enhances the model's ability to learn more robust and detailed feature representations.

\begin{figure}
    \centering
    \includegraphics[width=\linewidth]{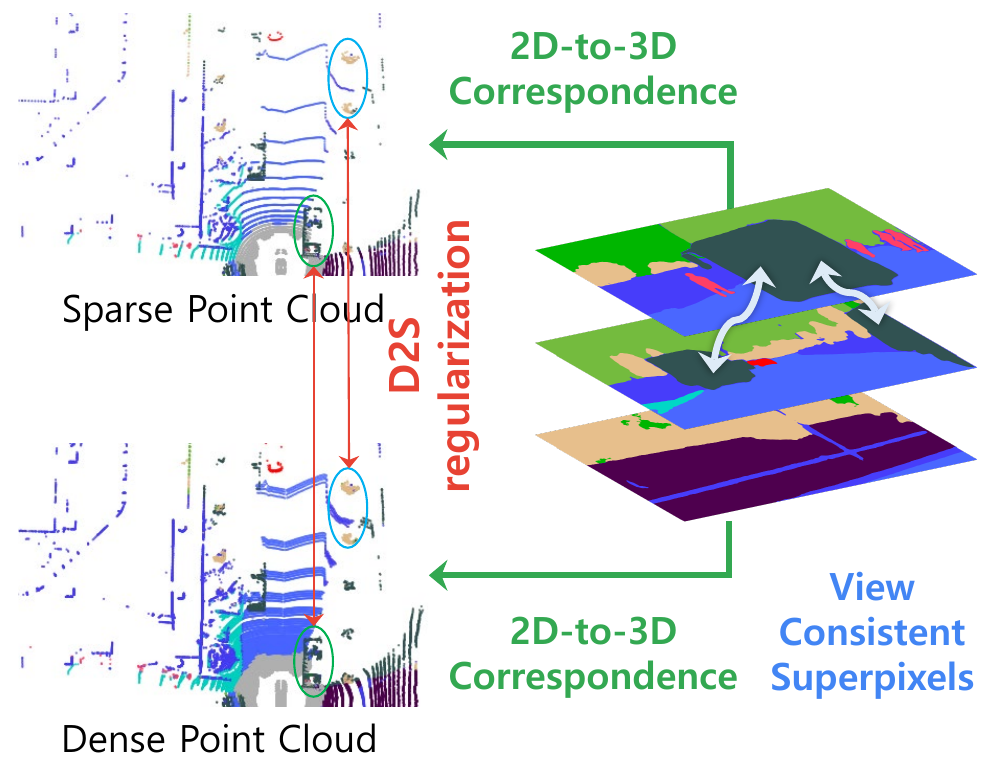}
    \vspace{-0.6cm}
    \caption{\textbf{Dense-to-sparse (D2S) Consistency Regularization Module.} Dense point clouds are generated by aggregating multi-sweep LiDAR scans captured over a defined time window. The D2S regularization enforces consistency between the dense and sparse point clouds, improving the model's ability to learn robust and detailed representations.}
    \label{fig:d2s}
\end{figure}

\begin{figure*}[t]
    \centering
    \includegraphics[width=\textwidth]{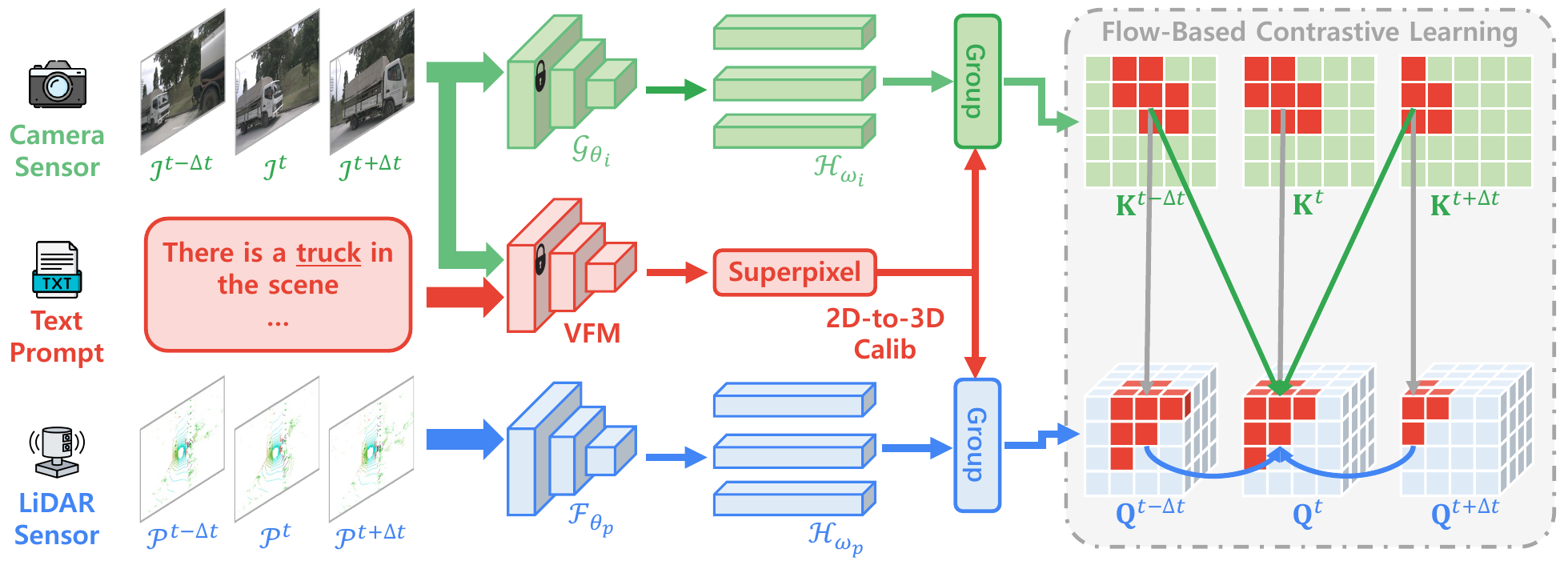}
    \vspace{-0.5cm}
    \caption{\textbf{Flow-based contrastive learning (FCL) pipeline.} The FCL pipeline processes multiple LiDAR-camera pairs captured over consecutive scans. It leverages temporally aligned semantic superpixels to define three contrastive learning objectives: (1) \textbf{Spatial Contrastive Learning}, which enforces feature consistency between LiDAR and camera modalities within the same frame, (2) \textbf{Intra-Sensor Temporal Contrastive Learning}, which aligns features across consecutive LiDAR scans, ensuring temporal coherence in dynamic scenes, and (3) \textbf{Cross-Sensor Temporal Contrastive Learning}, which aligns LiDAR features with temporally adjacent camera frames to enhance cross-modal consistency. This approach improves multi-modal representation learning by integrating spatial and temporal consistency.}
    \label{fig:framework}
\end{figure*}

\noindent\textbf{Point Cloud Concatenation.}
Given a keyframe point cloud $\mathcal{P}^t$ captured at time $t$ and a sequence of sweep point clouds $\{\mathcal{P}^s | s = 1, ..., T\}$ captured at earlier timestamps $s$, we align the coordinates $(x_i^s, y_i^s, z_i^s)$ of each sweep point cloud $\mathcal{P}^s$ with the coordinate system of $\mathcal{P}^t$. The transformation adjusts for vehicle motion and is defined as follows:
\begin{equation}
    [\widetilde{x}_i^s, \widetilde{y}_i^s, \widetilde{z}_i^s]^{\text{T}} = \Gamma_{s \to t} \times [x_i^s, y_i^s, z_i^s]^{\text{T}}~,
    \label{equ:sweep}
\end{equation}
where $\Gamma_{s \to t}$ denotes the transformation matrix that maps sweep point cloud coordinates from time $s$ to the keyframe coordinate system at time $t$. Next, we concatenate the transformed sweep points $\{\widetilde{\mathcal{P}}^s | s = 1, ..., T\}$ with the keyframe point cloud $\mathcal{P}^t$ to form a dense point cloud $\mathcal{P}^d$. As depicted in \cref{fig:d2s}, the resulting $\mathcal{P}^d$ combines temporal information from consecutive sweeps, providing a denser and enriched representation that supports enhanced feature learning.

\noindent\textbf{Dense Superpoints.}
To enrich point cloud representations, we generate superpoints $\mathcal{O}_p^d$ and $\mathcal{O}_p^t$ for the dense point cloud $\mathcal{P}^d$ and the keyframe point cloud $\mathcal{P}^t$, respectively, using their associated superpixels $\mathcal{O}_c^t$. Both $\mathcal{P}^t$ and $\mathcal{P}^d$ are processed through a shared-weight 3D network $\mathcal{F}_{\theta_p}$ and $\mathcal{H}_{\omega_p}$ for feature extraction. The extracted features are grouped by averaging based on their superpoint indices, resulting in superpoint feature representations $\mathbf{Q}^d$ and $\mathbf{Q}^t$, where $\mathbf{Q}^d \in \mathbb{R}^{M \times C}$ and $\mathbf{Q}^t \in \mathbb{R}^{M \times C}$. To enforce consistency in feature representations between the dense and sparse superpoints, we introduce the D2S loss as:
\begin{equation}
    \mathcal{L}_{\text{d2s}}(\mathbf{Q}^d, \mathbf{Q}^t) = \frac{1}{M}\sum_{i=1}^{M}(1 - <\mathbf{q}_i^d, \mathbf{q}_i^t>)~.
    \label{eq:regularization}
\end{equation}
This loss encourages the model to maintain consistent feature representations across varying point cloud densities, thereby improving robustness to density variations.

\subsection{FCL: Flow-Based Contrastive Learning}
\label{sec:spatial_temporal}

\noindent\textbf{Motivation.} LiDAR point clouds are inherently sequential, capturing dynamic scene information across consecutive timestamps. However, prior works \cite{sautier2022slidr,mahmoud2023st-slidr,liu2023seal} primarily focus on individual LiDAR scans, neglecting the temporal consistency of moving objects across sequential scenes. To address these limitations, we propose a flow-based contrastive learning framework that leverages sequential LiDAR data to enforce spatiotemporal consistency, ensuring robust feature representation across dynamic environments.

\noindent\textbf{Spatial Contrastive Learning.} Our framework, as illustrated in \cref{fig:framework}, processes three LiDAR-camera pairs from different timestamps within a suitable time window, specifically $\{(\mathcal{P}^{t}, \mathcal{I}^{t}), (\mathcal{P}^{t + \Delta t}, \mathcal{I}^{t + \Delta t}), (\mathcal{P}^{t - \Delta t}, \mathcal{I}^{t - \Delta t})\}$, where $t$ denotes the current timestamp, and $\Delta t$ represents the timespan. Similar to previous works \cite{sautier2022slidr,liu2023seal}, we distill knowledge from a 2D network into a 3D network for each scene individually. Taking the pair $(\mathcal{P}^{t}, \mathcal{I}^{t})$ as an example, the LiDAR point cloud $\mathcal{P}^{t}$ and the corresponding image $\mathcal{I}^{t}$ are first processed by the 3D and 2D networks to extract per-point and per-pixel features, respectively. These features are subsequently grouped via average pooling based on their respective superpoints $\mathcal{O}_{p}^{t}$ and superpixels $\mathcal{O}_{c}^{t}$ to yield superpoint features $\mathbf{Q}^{t}$ and superpixel features $\mathbf{K}^{t}$. To enforce alignment between the 3D and 2D representations, we propose a spatial contrastive loss that leverages pretrained 2D priors to guide 3D feature learning. The loss is defined as:
\begin{equation}
    \mathcal{L}_\text{sc}(\mathbf{Q}^t, \mathbf{K}^t) = - \frac{1}{M} \sum_{i=1}^{M} \log \left[ \frac{e^{(<\mathbf{q}_i^t, \mathbf{k}_i^t> / \tau)}}{\sum_{j=1}^M e^{(<\mathbf{q}_i^t, \mathbf{k}_j^t> / \tau)}} \right]~.
    \label{eq:spatial}
\end{equation}
This loss promotes strong feature correspondence between spatially aligned superpoints and superpixels while suppressing alignment with irrelevant pairs. The overall spatial contrastive loss is computed as the mean of $\mathcal{L}_\text{sc}(\mathbf{Q}^t, \mathbf{K}^t)$, $\mathcal{L}_\text{sc}(\mathbf{Q}^{t - \Delta t}, \mathbf{K}^{t - \Delta t})$, and $\mathcal{L}_\text{sc}(\mathbf{Q}^{t + \Delta t}, \mathbf{K}^{t + \Delta t})$.

\noindent\textbf{Intra-Sensor Temporal Contrastive Learning.}
While the spatial contrastive learning objective in \cref{eq:spatial} effectively aligns image and point cloud features within the same scene, it does not explicitly enforce consistency for moving objects across different scenes. To address this, we introduce a temporal contrastive loss, designed to ensure that dynamic objects maintain consistent feature representations across sequential LiDAR scans. Given two point cloud frames, $\mathcal{P}^{t}$ and $\mathcal{P}^{t + \Delta t}$, we extract their corresponding superpoint features, $\mathbf{Q}^{t}$ and $\mathbf{Q}^{t + \Delta t}$, by aggregating features within each superpoint. The temporal contrastive loss encourages alignment between these superpoints across timestamps:
\begin{equation}
    \mathcal{L}_{\text{tc}}(\mathbf{Q}^{t}, \mathbf{Q}^{t + \Delta t}) = - \frac{1}{M} \sum_{i=1}^{M} \log \left[ \frac{e^{(<\mathbf{q}_{i}^{t}, \mathbf{q}_{i}^{t + \Delta t}> / \tau)}}{\sum_{j=1}^M e^{(<\mathbf{q}_{i}^{t}, \mathbf{q}_{j}^{t + \Delta t}> / \tau)}} \right]~.
    \label{eq:temporal}
\end{equation}
The same temporal contrastive loss is applied between $\mathbf{Q}^{t}$ and $\mathbf{Q}^{t - \Delta t}$. The overall temporal contrastive loss is then computed as the mean of $\mathcal{L}_{\text{tc}}(\mathbf{Q}^{t}, \mathbf{Q}^{t + \Delta t})$ and $\mathcal{L}_{\text{tc}}(\mathbf{Q}^{t}, \mathbf{Q}^{t - \Delta t})$. This mechanism enriches the feature representations at time $t$ with context-aware information from both preceding and succeeding scenes.

\noindent\textbf{Cross-Sensor Temporal Contrastive Learning.}
To further enhance temporal consistency in scenarios involving multi-sensors, we introduce a cross-sensor temporal contrastive loss. This loss aligns features extracted from different sensors at adjacent timestamps, ensuring robust cross-sensor temporal coherence. For example, the cross-sensor temporal contrastive loss between superpoint features $\mathbf{Q}^{t}$ and superpixel features $\mathbf{K}^{t + \Delta t}$ is defined as:
\begin{equation}
    \mathcal{L}_{\text{cc}}(\mathbf{Q}^{t}, \mathbf{K}^{t + \Delta t}) = - \frac{1}{M} \sum_{i=1}^{M} \log \left[ \frac{e^{(<\mathbf{q}_{i}^{t}, \mathbf{k}_{i}^{t + \Delta t}> / \tau)}}{\sum_{j=1}^M e^{(<\mathbf{q}_{i}^{t}, \mathbf{k}_{j}^{t + \Delta t}> / \tau)}} \right]~.
    \label{eq:cross-temporal}
\end{equation}
This cross-sensor term is similarly computed between $\mathbf{Q}^{t}$ and $\mathbf{K}^{t - \Delta t}$. By incorporating both single-sensor and cross-sensor temporal contrastive loss, the proposed framework enhances the robustness of learned representations across diverse viewpoints and sensor modalities, making it more adaptable to real-world scenarios.

\subsection{Temporal Voting Semantic Segmentation}
\label{sec:temporal_voting}

\noindent\textbf{Motivation.} LiDAR point clouds inherently capture sequential data. However, existing downstream segmentation methods \cite{sautier2022slidr,liu2023seal,xu2024superflow} typically predict semantic labels frame by frame, overlooking semantic coherence across consecutive scenes. To address this limitation, we propose a temporal voting segmentation strategy that refines semantic predictions for the current frame by incorporating predictions from neighboring frames. This approach ensures consistent and coherent predictions across sequential scenes.

\noindent\textbf{Temporal Voting Segmentation.} Consider three consecutive LiDAR point clouds, $\{\mathcal{P}^{t-1}, \mathcal{P}^{t},\mathcal{P}^{t+1}\}$. A 3D network independently generates semantic scores for each point cloud, yielding $\{\mathbf{P}^{t-1},\mathbf{P}^{t},\mathbf{P}^{t+1}\}$. To refine the semantic probabilities $\mathbf{P}^{t}$, we propose a temporal voting strategy that integrates information from $\mathbf{P}^{t-1}$ and $\mathbf{P}^{t+1}$, promoting temporal semantic coherence across the sequence. Specifically, the point clouds are first transformed into a unified coordinate system (\textit{e.g.}, the world coordinate system or the coordinate system of the $t$-th LiDAR frame), resulting in $\{\widehat{\mathcal{P}}^{t-1},\widehat{\mathcal{P}}^{t},\widehat{\mathcal{P}}^{t+1}\}$. This transformation ensures spatial alignment, enabling direct comparisons across frames. For each point in $\widehat{\mathcal{P}}^{t}$, we identify its nearest neighbors in $\widehat{\mathcal{P}}^{t-1}$ and $\widehat{\mathcal{P}}^{t+1}$ using the Euclidean distance. A distance threshold $\sigma$ is introduced to filter neighbors: if the distance from a point in $\widehat{\mathcal{P}}^{t}$ to its nearest neighbor in $\widehat{\mathcal{P}}^{t-1}$ or $\widehat{\mathcal{P}}^{t+1}$ is less than $\sigma$, the semantic probability of the neighbor is interpolated to the current point. This process blends semantic information from adjacent frames, refining the predictions for $\mathcal{P}^{t}$. The final semantic probabilities $\widehat{\mathbf{P}}^{t}$ are obtained by averaging the interpolated probabilities with the network's direct predictions, balancing temporal information and local frame accuracy. By leveraging temporal voting, this method enhances semantic consistency across consecutive scenes, addressing challenges posed by dynamic environments and moving objects. The detailed pseudo-code for this process is provided in \cref{alg:temporal_voting}.

\begin{algorithm}[!t]
    \caption{Temporal Voting Semantic Segmentation}
    \label{alg:temporal_voting}
    \begin{algorithmic}[1]
        \footnotesize
        \STATE {\bfseries Input:} Point clouds $\mathcal{P}^{t-1} \in \mathbb{R}^{N_{t-1} \times (3+L)}$, $\mathcal{P}^{t} \in \mathbb{R}^{N_{t} \times (3+L)}$, $\mathcal{P}^{t+1} \in \mathbb{R}^{N_{t+1} \times (3+L)}$ from three consecutive scans, 3D network $\mathcal{F}_{\theta_{p}}$, a distance threshold $\sigma$, transformation matrices $T_{t-1}$, $T_{t}$, $T_{t+1}$ for unifying coordinate systems.
        \STATE {\bfseries Output:} Predicted semantic labels $y^{t}$ for $\mathcal{P}^t$.
        \STATE Compute 3D semantic scores for each point cloud: $\mathbf{P}^{t-1} = \mathcal{F}_{\theta_{p}}(\mathcal{P}^{t-1})$, $\mathbf{P}^{t} = \mathcal{F}_{\theta_{p}}(\mathcal{P}^{t})$, $\mathbf{P}^{t+1} = \mathcal{F}_{\theta_{p}}(\mathcal{P}^{t+1})$.
        \STATE Transform point clouds into a unified system: $\widehat{\mathcal{P}}^{t-1} = T_{t-1}(\mathcal{P}^{t-1})$, $\widehat{\mathcal{P}}^{t} = T_{t}(\mathcal{P}^{t})$, $\widehat{\mathcal{P}}^{t+1} = T_{t+1}(\mathcal{P}^{t+1})$.
        \FOR{$i = 1$ \TO $N_{t}$}
            \STATE Initialize: $n = 1$.
            \STATE Compute the nearest distance and index:
                \vspace{-0.2cm}
                \begin{align*}
                    d_{t-1}, j_{t-1} &= \min(\|\widehat{\mathcal{P}}^{t}[i] - \widehat{\mathcal{P}}^{t-1}\|_2), \\
                    d_{t+1}, j_{t+1} &= \min(\|\widehat{\mathcal{P}}^{t}[i] - \widehat{\mathcal{P}}^{t+1}\|_2).
                \end{align*}
            \vspace{-0.4cm}
            \IF{$d_{t-1} < \sigma$}
                \STATE $\mathbf{P}^{t}[i] \gets \mathbf{P}^{t}[i] + \mathbf{P}^{t-1}[j_{t-1}]$, $n \gets n + 1$.
            \ENDIF
            \IF{$d_{t+1} < \sigma$}
                \STATE $\mathbf{P}^{t}[i] \gets \mathbf{P}^{t}[i] + \mathbf{P}^{t+1}[j_{t+1}]$, $n \gets n + 1$.
            \ENDIF
            \STATE Average: $\mathbf{P}^{t}[i] \gets \mathbf{P}^{t}[i] / n$.
        \ENDFOR
        \STATE Predict semantic labels: $y^{t} = \text{argmax}(\mathbf{P}^{t}, \text{dim}=1)$.
        \RETURN $y^{t}$.
    \end{algorithmic}
\end{algorithm}

\section{Experiments}
\label{sec:experiments}
This section presents a comprehensive evaluation of the proposed SuperFlow++ framework. We begin by detailing the experimental setup, including datasets, metrics, and implementation details (\cref{sec:settings}). Subsequently, we perform a comparative analysis to showcase the superiority of our approach over state-of-the-art methods across diverse tasks and datasets (\cref{sec:comparative_study}). Lastly, we conduct an in-depth ablation study to examine the contribution of each component in our framework and its impact on overall performance (\cref{sec:ablation_study}).

\begin{table*}[t]
    \centering
    \caption{\textbf{Comparisons of state-of-the-art pretraining methods} pretrained on \textit{nuScenes} \cite{fong2022Panoptic-nuScenes} and fine-tuned on \textit{nuScenes} \cite{fong2022Panoptic-nuScenes}, \textit{SemanticKITTI} \cite{behley2019SemanticKITTI} and \textit{Waymo Open} \cite{sun2020waymoOpen} with specified data portions, respectively. All methods use MinkUNet \cite{choy2019minkowski} as the 3D semantic segmentation backbone. \textbf{LP} denotes linear probing with a frozen backbone. All scores are given in percentage (\%). The \textbf{best} and \underline{second best} scores in each configuration are highlighted in \textbf{bold} and \underline{underline}, respectively.}
    \vspace{-0.2cm}
    \label{tab:benchmark}
    \resizebox{\textwidth}{!}{
    \begin{tabular}{r|r|p{52pt}<{\centering}|p{52pt}<{\centering}|p{18pt}<{\centering}p{18pt}<{\centering}p{18pt}<{\centering}p{18pt}<{\centering}p{18pt}<{\centering}p{18pt}<{\centering}|p{20pt}<{\centering}|p{20pt}<{\centering}}
        \toprule
        \multirow{2}{*}{\textbf{Method}} & \multirow{2}{*}{\textbf{Venue}} & \textbf{Backbone}  & \textbf{Backbone} & \multicolumn{6}{c}{\textbf{nuScenes}} \vline & \textbf{KITTI} & \textbf{Waymo}
        \\
        & & \textbf{(2D)} & \textbf{(3D)} & \textbf{LP} & \textbf{1\%} & \textbf{5\%} & \textbf{10\%} & \textbf{25\%} & \textbf{Full} & \textbf{1\%} & \textbf{1\%}
        \\\midrule\midrule
        \cellcolor{sf_gray!18}Random & \cellcolor{sf_gray!18}- & \cellcolor{sf_gray!18}None & \cellcolor{sf_gray!18}MinkUNet-34 & \cellcolor{sf_gray!18}8.10 & \cellcolor{sf_gray!18}30.30 & \cellcolor{sf_gray!18}47.84 & \cellcolor{sf_gray!18}56.15 & \cellcolor{sf_gray!18}65.48 & \cellcolor{sf_gray!18}74.66 & \cellcolor{sf_gray!18}39.50 & \cellcolor{sf_gray!18}39.41
        \\\midrule
        PointContrast \cite{xie2020pointcontrast} & ECCV'20 & \multirow{4}{*}{None} & \multirow{4}{*}{MinkUNet-34} & \underline{21.90} & 32.50 & - & - & - & - & \underline{41.10} & -
        \\
        DepthContrast \cite{zhang2021depthcontrast} & ICCV'21 & & & \textbf{22.10} & 31.70 & - & - & - & - & \textbf{41.50} & -
        \\
        ALSO \cite{boulch2023also} & CVPR'23 & & & - & \underline{37.70} & - & \underline{59.40} & - & \underline{72.00} & - & -
        \\
        BEVContrast \cite{sautier2024bevcontrast} & 3DV'24 & & & - & \textbf{38.30} & - & \textbf{59.60} & - & \textbf{72.30} & - & -
        \\\midrule
        PPKT \cite{liu2021ppkt} & arXiv'21 & \multirow{9}{*}{\makecell{ResNet-50 \\ \cite{he2016resnet}}} & \multirow{9}{*}{\makecell{MinkUNet-34 \\ \cite{choy2019minkowski}}} & 35.90 & 37.80 & 53.74 & 60.25 & 67.14 & 74.52 & 44.00 & 47.60
        \\
        SLidR \cite{sautier2022slidr} & CVPR'22 & & & 38.80 & 38.30 & 52.49 & 59.84 & 66.91 & 74.79 & 44.60 & 47.12
        \\
        ST-SLidR \cite{mahmoud2023st-slidr} & CVPR'23 & & & 40.48 & 40.75 & 54.69 & 60.75 & 67.70 & 75.14 & 44.72 & 44.93
        \\
        TriCC \cite{pang2023tricc} & CVPR'23 & & & 38.00 & 41.20 & 54.10 & 60.40 & 67.60 & 75.60 & 45.90 & -
        \\
        Seal \cite{liu2023seal} & NeurIPS'23 & & & 44.95 & 45.84 & 55.64 & 62.97 & 68.41 & 75.60 & 46.63 & 49.34
        \\
        CSC \cite{chen2024csc} & CVPR'24 & & & \underline{46.00} & \underline{47.00} & \underline{57.00} & \underline{63.30} & 68.60 & 75.70 & 47.20 & -
        \\
        HVDistill \cite{zhang2024hvdistill} & IJCV'24 & & & 39.50 & 42.70 & 56.60 & 62.90 & \underline{69.30} & \underline{76.60} & \underline{49.70} & -
        \\
        SuperFlow \cite{xu2024superflow} & ECCV'24 & & & 45.27 & 46.79 & 56.93 & 63.03 & 68.94 & 76.23 & 48.84 & \underline{49.78}
        \\
        \cellcolor{sf_blue!8}\textbf{SuperFlow++} & \cellcolor{sf_blue!8}\textbf{Ours} & & & \cellcolor{sf_blue!8}\textbf{46.45} & \cellcolor{sf_blue!8}\textbf{47.62} & \cellcolor{sf_blue!8}\textbf{57.25} & \cellcolor{sf_blue!8}\textbf{63.61} & \cellcolor{sf_blue!8}\textbf{69.41} & \cellcolor{sf_blue!8}\textbf{76.62} & \cellcolor{sf_blue!8}\underline{49.53} & \cellcolor{sf_blue!8}\textbf{50.65}
        \\\midrule
        PPKT \cite{liu2021ppkt} & arXiv'21 & \multirow{8}{*}{\makecell{ViT-S \\ \cite{oquab2023dinov2}}} & \multirow{8}{*}{\makecell{MinkUNet-34 \\ \cite{choy2019minkowski}}} & 38.60 & 40.60 & 52.06 & 59.99 & 65.76 & 73.97 & 43.25 & 47.44
        \\
        SLidR \cite{sautier2022slidr} & CVPR'22 & & & 44.70 & 41.16 & 53.65 & 61.47 & 66.71 & 74.20 & 44.67 & 47.57
        \\
        Seal \cite{liu2023seal} & NeurIPS'23 & & & 45.16 & 44.27 & 55.13 & 62.46 & 67.64 & 75.58 & 46.51 & 48.67
        \\
        ScaLR \cite{puy2024scalar} & CVPR'24 & & & 42.40 & 40.50 & - & - & - & - & - & -
        \\
        SuperFlow \cite{xu2024superflow} & ECCV'24 & & & 46.44 & \underline{47.81} & \underline{59.44} & \underline{64.47} & 69.20 & \underline{76.54} & 47.97 & 49.94
        \\
        NCLR \cite{zhang2025nclr} & TPAMI'25 & & & 42.81 & 42.29 & - & 59.50 & \textbf{71.20} & 72.70 & 45.84 & -
        \\
        LargeAD \cite{kong2025largead} & TPAMI'25 & & & \underline{46.58} & 46.78 & 57.33 & 63.85 & 68.66 & 75.75 & \textbf{50.07} & \underline{50.83}
        \\
        \cellcolor{sf_red!8}\textbf{SuperFlow++~~} & \cellcolor{sf_red!8}\textbf{Ours} & & & \cellcolor{sf_red!8}\textbf{48.57} & \cellcolor{sf_red!8}\textbf{49.07} & \cellcolor{sf_red!8}\textbf{60.57} & \cellcolor{sf_red!8}\textbf{65.21} & \cellcolor{sf_red!8}\underline{70.05} & \cellcolor{sf_red!8}\textbf{76.92} & \cellcolor{sf_red!8}\underline{49.27} & \cellcolor{sf_red!8}\textbf{51.25}
        \\\midrule
        PPKT \cite{liu2021ppkt} & arXiv'21 & \multirow{5}{*}{\makecell{ViT-B \\ \cite{oquab2023dinov2}}} & \multirow{5}{*}{\makecell{MinkUNet-34 \\ \cite{choy2019minkowski}}} & 39.95 & 40.91 & 53.21 & 60.87 & 66.22 & 74.07 & 44.09 & 47.57
        \\
        SLidR \cite{sautier2022slidr} & CVPR'22 & & & 45.35 & 41.64 & 55.83 & 62.68 & 67.61 & 74.98 & 45.50 & 48.32
        \\
        Seal \cite{liu2023seal} & NeurIPS'23 & & & 46.59 & 45.98 & 57.15 & 62.79 & 68.18 & 75.41 & 47.24 & 48.91
        \\
        SuperFlow \cite{xu2024superflow} & ECCV'24 & & & \underline{47.66} & \underline{48.09} & \underline{59.66} & \underline{64.52} & \underline{69.79} & \underline{76.57} & \underline{48.40} & \underline{50.20}
        \\
        \cellcolor{sf_yellow!8}\textbf{SuperFlow++~~} & \cellcolor{sf_yellow!8}\textbf{Ours} & & & \cellcolor{sf_yellow!8}\textbf{48.86} & \cellcolor{sf_yellow!8}\textbf{49.56} & \cellcolor{sf_yellow!8}\textbf{60.75} & \cellcolor{sf_yellow!8}\textbf{65.46} & \cellcolor{sf_yellow!8}\textbf{70.19} & \cellcolor{sf_yellow!8}\textbf{77.29} & \cellcolor{sf_yellow!8}\textbf{49.90} & \cellcolor{sf_yellow!8}\textbf{51.65}
        \\\midrule
        PPKT \cite{liu2021ppkt} & arXiv'21 & \multirow{5}{*}{\makecell{ViT-L \\ \cite{oquab2023dinov2}}} & \multirow{5}{*}{\makecell{MinkUNet-34 \\ \cite{choy2019minkowski}}} & 41.57 & 42.05 & 55.75 & 61.26 & 66.88 & 74.33 & 45.87 & 47.82
        \\
        SLidR \cite{sautier2022slidr} & CVPR'22 & & & 45.70 & 42.77 & 57.45 & 63.20 & 68.13 & 75.51 & 47.01 & 48.60
        \\
        Seal \cite{liu2023seal} & NeurIPS'23 & & & 46.81 & 46.27 & 58.14 & 63.27 & 68.67 & 75.66 & 47.55 & 50.02
        \\
        SuperFlow \cite{xu2024superflow} & ECCV'24 & & & \underline{48.01} & \underline{49.95} & \underline{60.72} & \underline{65.09} & \underline{70.01} & \underline{77.19} & \underline{49.07} & \underline{50.67}
        \\
        \cellcolor{sf_green!8}\textbf{SuperFlow++~~} & \cellcolor{sf_green!8}\textbf{Ours} & & & \cellcolor{sf_green!8}\textbf{49.78} & \cellcolor{sf_green!8}\textbf{50.92} & \cellcolor{sf_green!8}\textbf{61.83} & \cellcolor{sf_green!8}\textbf{66.30} & \cellcolor{sf_green!8}\textbf{71.07} & \cellcolor{sf_green!8}\textbf{77.63} & \cellcolor{sf_green!8}\textbf{50.33} & \cellcolor{sf_green!8}\textbf{52.12}
        \\\bottomrule
    \end{tabular}}
\end{table*}

\begin{table*}[t]
    \centering
    \caption{\textbf{Domain generalization study} of different pretraining methods pretrained on the \textit{nuScenes} \cite{caesar2020nuScenes} dataset and fine-tuned on other \textit{seven} heterogeneous 3D semantic segmentation datasets with specified data portions, respectively. All scores are given in percentage (\%). The best scores in each configuration are highlighted in \textbf{bold}.}
    \vspace{-0.2cm}
    \label{tab:multiple_datasets}
    \resizebox{\textwidth}{!}{
    \begin{tabular}{r|r|cc|cc|cc|cc|cc|cc|cc}
        \toprule
        \multirow{2}{*}{\textbf{Method}} & \multirow{2}{*}{\textbf{Venue}} & \multicolumn{2}{c}{\textbf{ScriKITTI}} \vline & \multicolumn{2}{c}{\textbf{Rellis-3D}} \vline & \multicolumn{2}{c}{\textbf{SemPOSS}} \vline & \multicolumn{2}{c}{\textbf{SemSTF}} \vline & \multicolumn{2}{c}{\textbf{SynLiDAR}} \vline & \multicolumn{2}{c}{\textbf{DAPS-3D}} \vline & \multicolumn{2}{c}{\textbf{Synth4D}}
        \\
        & & \textbf{1\%} & \textbf{10\%} & \textbf{1\%} & \textbf{10\%} & \textbf{Half} & \textbf{Full} & \textbf{Half} & \textbf{Full} & \textbf{1\%} & \textbf{10\%} & \textbf{Half} & \textbf{Full} & \textbf{1\%} & \textbf{10\%}
        \\\midrule\midrule
        \cellcolor{sf_gray!18}Random & \cellcolor{sf_gray!18}- & \cellcolor{sf_gray!18}23.81 & \cellcolor{sf_gray!18}47.60 & \cellcolor{sf_gray!18}38.46 & \cellcolor{sf_gray!18}53.60 & \cellcolor{sf_gray!18}46.26 & \cellcolor{sf_gray!18}54.12 & \cellcolor{sf_gray!18}48.03 & \cellcolor{sf_gray!18}48.15 & \cellcolor{sf_gray!18}19.89 & \cellcolor{sf_gray!18}44.74 & \cellcolor{sf_gray!18}74.32 & \cellcolor{sf_gray!18}79.38 & \cellcolor{sf_gray!18}20.22 & \cellcolor{sf_gray!18}66.87
        \\\midrule
        PPKT \cite{liu2021ppkt} & arXiv'21 & 36.50 & 51.67 & 49.71 & 54.33 & 50.18 & 56.00 & 50.92 & 54.69 & 37.57 & 46.48 & 78.90 & 84.00 & 61.10 & 62.41
        \\
        SLidR \cite{sautier2022slidr} & CVPR'22 & 39.60 & 50.45 & 49.75 & 54.57 & 51.56 & 55.36 & 52.01 & 54.35 & 42.05 & 47.84 & 81.00 & 85.40 & 63.10 & 62.67
        \\
        Seal \cite{liu2023seal} & NeurIPS'23 & 40.64 & 52.77 & 51.09 & 55.03 & 53.26 & 56.89 & 53.46 & 55.36 & 43.58 & 49.26 & 81.88 & 85.90 & 64.50 & 66.96
        \\
        SuperFlow \cite{xu2024superflow} & ECCV'24 & \underline{42.70} & \underline{54.00} & \underline{52.83} & \underline{55.71} & \underline{54.41} & \underline{57.33} & \underline{54.72} & \underline{56.57} & \underline{44.85} & \underline{51.38} & \underline{82.43} & \underline{86.21} & \underline{65.31} & \underline{69.43}
        \\
        \cellcolor{sf_blue!8}\textbf{SuperFlow++~~} & \cellcolor{sf_blue!8}\textbf{Ours} & \cellcolor{sf_blue!8}\textbf{44.74} & \cellcolor{sf_blue!8}\textbf{56.03} & \cellcolor{sf_blue!8}\textbf{54.09} & \cellcolor{sf_blue!8}\textbf{56.82} & \cellcolor{sf_blue!8}\textbf{56.15} & \cellcolor{sf_blue!8}\textbf{57.13} & \cellcolor{sf_blue!8}\textbf{56.75} & \cellcolor{sf_blue!8}\textbf{58.10} & \cellcolor{sf_blue!8}\textbf{46.48} & \cellcolor{sf_blue!8}\textbf{52.71} & \cellcolor{sf_blue!8}\textbf{82.94} & \cellcolor{sf_blue!8}\textbf{86.78} & \cellcolor{sf_blue!8}\textbf{67.76} & \cellcolor{sf_blue!8}\textbf{71.12}
        \\\bottomrule
    \end{tabular}}
\end{table*}

\begin{table}[t]
    \centering
    \caption{Performance comparison of state-of-the-art pretraining methods pretrained and fine-tuned on the \textit{nuScenes} dataset \cite{caesar2020nuScenes}, using specified data proportions. All methods utilize CenterPoint \cite{yin2021centerpoint} and SECOND \cite{yan2018second} as 3D object detection backbones. All scores are given in percentage (\%). The best scores are highlighted in \textbf{bold}.}
    \vspace{-0.2cm}
    \label{tab:detection}
    \resizebox{\linewidth}{!}{
    \begin{tabular}{r|p{0.6cm}<{\centering}p{0.6cm}<{\centering}|p{0.6cm}<{\centering}p{0.6cm}<{\centering}|p{0.6cm}<{\centering}p{0.6cm}<{\centering}}
        \toprule
        \multirow{3.7}{*}{\textbf{Method}} & \multicolumn{6}{c}{\textbf{nuScenes}}
        \\\cmidrule{2-7}
        & \multicolumn{2}{c|}{\textbf{5\%}} & \multicolumn{2}{c|}{\textbf{10\%}} & \multicolumn{2}{c}{\textbf{20\%}}
        \\
        & \textbf{mAP} & \textbf{NDS} & \textbf{mAP} & \textbf{NDS} & \textbf{mAP} & \textbf{NDS}
        \\\midrule\midrule
        \multicolumn{7}{c}{\textit{Backbone: VoxelNet + CenterPoint}}
        \\\midrule
        \cellcolor{sf_gray!18}Random & \cellcolor{sf_gray!18}38.0 & \cellcolor{sf_gray!18}44.3 & \cellcolor{sf_gray!18}46.9 & \cellcolor{sf_gray!18}55.5 & \cellcolor{sf_gray!18}50.2 & \cellcolor{sf_gray!18}59.7
        \\
        PointContrast \cite{xie2020pointcontrast} & 39.8 & 45.1 & 47.7 & 56.0 & - & -
        \\
        GCC-3D \cite{liang2021exploring} & 41.1 & 46.8 & 48.4 & 56.7 & - & -
        \\
        SLidR \cite{sautier2022slidr} & 43.3 & 52.4 & 47.5 & 56.8 & 50.4 & 59.9
        \\
        TriCC \cite{pang2023tricc} & 44.6 & 54.4 & 48.9 & 58.1 & 50.9 & 60.3
        \\
        CSC \cite{chen2024csc} & 45.3 & 54.2 & 49.3 & 58.3 & 51.9 & 61.3
        \\
        SuperFlow \cite{xu2024superflow} & 46.0 & 54.9 & 49.7 & 58.5 & 52.5 & 61.5
        \\
        \cellcolor{sf_blue!8}\textbf{SuperFlow++} & \cellcolor{sf_blue!8}\textbf{47.1} & \cellcolor{sf_blue!8}\textbf{55.2} & \cellcolor{sf_blue!8}\textbf{50.3} & \cellcolor{sf_blue!8}\textbf{58.9} & \cellcolor{sf_blue!8}\textbf{52.9} & \cellcolor{sf_blue!8}\textbf{61.7}
        \\\midrule\midrule
        \multicolumn{7}{c}{\textit{Backbone: VoxelNet + SECOND}}
        \\\midrule
        \cellcolor{sf_gray!18}Random & \cellcolor{sf_gray!18}35.8 & \cellcolor{sf_gray!18}45.9 & \cellcolor{sf_gray!18}39.0 & \cellcolor{sf_gray!18}51.2 & \cellcolor{sf_gray!18}43.1 & \cellcolor{sf_gray!18}55.7
        \\
        SLidR \cite{sautier2022slidr} & 36.6 & 48.1 & 39.8 & 52.1 & 44.2 & 56.3
        \\
        TriCC \cite{pang2023tricc} & 37.8 & 50.0 & 41.4 & 53.5 & 45.5 & 57.7
        \\
        CSC \cite{chen2024csc} & 38.2 & 49.4 & 42.5 & 54.8 & 45.6 & 58.1
        \\
        SuperFlow \cite{xu2024superflow} & 38.3 & 49.8 & 42.7 & 55.2 & 45.8 & 58.5
        \\
        \cellcolor{sf_blue!8}\textbf{SuperFlow++} & \cellcolor{sf_blue!8}\textbf{39.1} & \cellcolor{sf_blue!8}\textbf{50.4} & \cellcolor{sf_blue!8}\textbf{43.2} & \cellcolor{sf_blue!8}\textbf{56.0} & \cellcolor{sf_blue!8}\textbf{46.2} & \cellcolor{sf_blue!8}\textbf{58.8}
        \\\bottomrule
    \end{tabular}}
\end{table}

\begin{table*}[t]
    \centering
    \caption{\textbf{Out-of-distribution 3D robustness study} of state-of-the-art pretraining methods under corruption and sensor failure scenarios in the \textit{nuScenes-C} dataset from the \textit{Robo3D} benchmark \cite{kong2023robo3D}. \textbf{Full} denotes fine-tuning with full labels. \textbf{LP} denotes linear probing with a frozen backbone. All mCE ($\downarrow$), mRR ($\uparrow$), and mIoU ($\uparrow$) scores are given in percentage (\%). The best scores in each configuration are highlighted in \textbf{bold}.}
    \vspace{-0.2cm}
    \label{tab:robo3d}
    \resizebox{\textwidth}{!}{
        \begin{tabular}{c|r|r|p{26pt}<{\centering}|p{26pt}<{\centering}|p{21pt}<{\centering}p{21pt}<{\centering}p{21pt}<{\centering}p{21pt}<{\centering}p{21pt}<{\centering}p{21pt}<{\centering}p{21pt}<{\centering}c|c}
        \toprule
        \textbf{\#} & \textbf{Initial} & \textbf{Backbone} & \textbf{mCE} & \textbf{mRR} & \textbf{Fog} & \textbf{Rain} & \textbf{Snow} & \textbf{Blur} & \textbf{Beam} & \textbf{Cross} & \textbf{Echo} & \textbf{Sensor} & \textbf{Average}
        \\\midrule\midrule
        \multirow{17}{*}{\rotatebox[origin=c]{90}{\textbf{Full}}} & \cellcolor{sf_gray!18}Random & \cellcolor{sf_gray!18}MinkU-18 & \cellcolor{sf_gray!18}115.61 & \cellcolor{sf_gray!18}70.85 & \cellcolor{sf_gray!18}53.90 & \cellcolor{sf_gray!18}71.10 & \cellcolor{sf_gray!18}48.22 & \cellcolor{sf_gray!18}51.85 & \cellcolor{sf_gray!18}62.21 & \cellcolor{sf_gray!18}37.73 & \cellcolor{sf_gray!18}57.47 & \cellcolor{sf_gray!18}38.97 & \cellcolor{sf_gray!18}52.68
        \\
        & SuperFlow \cite{xu2024superflow} & MinkU-18 & 109.00 & 75.66 & 54.95 & 72.79 & 49.56 & 57.68 & 62.82 & 42.45 & 59.61 & 41.77 & 55.21
        \\
        & \cellcolor{sf_green!8}\textbf{SuperFlow++} & \cellcolor{sf_green!8} MinkU-18 & \cellcolor{sf_green!8}\textbf{105.96} & \cellcolor{sf_green!8}\textbf{76.12} & \cellcolor{sf_green!8}\textbf{56.83} & \cellcolor{sf_green!8}\textbf{73.13} & \cellcolor{sf_green!8}\textbf{51.24} & \cellcolor{sf_green!8}\textbf{59.10} & \cellcolor{sf_green!8}\textbf{63.62} & \cellcolor{sf_green!8}\textbf{43.34} & \cellcolor{sf_green!8}\textbf{61.24} & \cellcolor{sf_green!8}\textbf{43.03} & \cellcolor{sf_green!8}\textbf{56.44}
        \\\cmidrule{2-14}
        & \cellcolor{sf_gray!18}Random & \cellcolor{sf_gray!18}MinkU-34 & \cellcolor{sf_gray!18}112.20 & \cellcolor{sf_gray!18}72.57 & \cellcolor{sf_gray!18}62.96 & \cellcolor{sf_gray!18}70.65 & \cellcolor{sf_gray!18}55.48 & \cellcolor{sf_gray!18}51.71 & \cellcolor{sf_gray!18}62.01 & \cellcolor{sf_gray!18}31.56 & \cellcolor{sf_gray!18}59.64 & \cellcolor{sf_gray!18}39.41 & \cellcolor{sf_gray!18}54.18
        \\
        & PPKT \cite{liu2021ppkt} & MinkU-34 & 105.64 & 75.87 & 64.01 & 72.18 & 59.08 & 57.17 & 63.88 & 36.34 & 60.59 & 39.57 & 56.60
        \\
        & SLidR \cite{sautier2022slidr} & MinkU-34 & 106.08 & 75.99 & 65.41 & 72.31 & 56.01 & 56.07 & 62.87 & 41.94 & 61.16 & 38.90 & 56.83
        \\
        & Seal \cite{liu2023seal} & MinkU-34 & 92.63 & 83.08 & \textbf{72.66} & 74.31 & \textbf{66.22} & \textbf{66.14} & 65.96 & 57.44 & \textbf{59.87} & 39.85 & 62.81
        \\
        & SuperFlow \cite{xu2024superflow} & MinkU-34 & 91.67 & 83.17 & 70.32 & 75.77 & 65.41 & 61.05 & 68.09 & 60.02 & 58.36 & 50.41 & 63.68
        \\
        & \cellcolor{sf_blue!8}\textbf{SuperFlow++} & \cellcolor{sf_blue!8} MinkU-34 & \cellcolor{sf_blue!8}\textbf{89.65} & \cellcolor{sf_blue!8}\textbf{83.65} & \cellcolor{sf_blue!8}72.15 & \cellcolor{sf_blue!8}\textbf{76.02} & \cellcolor{sf_blue!8}65.50 & \cellcolor{sf_blue!8}61.21 & \cellcolor{sf_blue!8}\textbf{69.31} & \cellcolor{sf_blue!8}\textbf{63.15} & \cellcolor{sf_blue!8}58.45 & \cellcolor{sf_blue!8}\textbf{51.43} & \cellcolor{sf_blue!8}\textbf{65.03}
        \\\cmidrule{2-14}
        & \cellcolor{sf_gray!18}Random & \cellcolor{sf_gray!18}MinkU-50 & \cellcolor{sf_gray!18}113.76 & \cellcolor{sf_gray!18}72.81 & \cellcolor{sf_gray!18}49.95 & \cellcolor{sf_gray!18}71.16 & \cellcolor{sf_gray!18}45.36 & \cellcolor{sf_gray!18}55.55 & \cellcolor{sf_gray!18}62.84 & \cellcolor{sf_gray!18}36.94 & \cellcolor{sf_gray!18}59.12 & \cellcolor{sf_gray!18}43.15 & \cellcolor{sf_gray!18}53.01
        \\
        & SuperFlow \cite{xu2024superflow} & MinkU-50 & 107.35 & 74.02 & 54.36 & 73.08 & 50.07 & 56.92 & 64.05 & 38.10 & 62.02 & 47.02 & 55.70
        \\
        & \cellcolor{sf_red!8}\textbf{SuperFlow++} & \cellcolor{sf_red!8} MinkU-50 & \cellcolor{sf_red!8}\textbf{103.11} & \cellcolor{sf_red!8}\textbf{76.04} & \cellcolor{sf_red!8}\textbf{55.42} & \cellcolor{sf_red!8}\textbf{74.20} & \cellcolor{sf_red!8}\textbf{52.96} & \cellcolor{sf_red!8}\textbf{57.30} & \cellcolor{sf_red!8}\textbf{66.49} & \cellcolor{sf_red!8}\textbf{39.37} & \cellcolor{sf_red!8}\textbf{63.79} & \cellcolor{sf_red!8}\textbf{50.32} & \cellcolor{sf_red!8}\textbf{57.48}
        \\\cmidrule{2-14}
        & \cellcolor{sf_gray!18}Random & \cellcolor{sf_gray!18}MinkU-101 & \cellcolor{sf_gray!18}109.10 & \cellcolor{sf_gray!18}74.07 & \cellcolor{sf_gray!18}50.45 & \cellcolor{sf_gray!18}73.02 & \cellcolor{sf_gray!18}48.85 & \cellcolor{sf_gray!18}58.48 & \cellcolor{sf_gray!18}64.18 & \cellcolor{sf_gray!18}43.86 & \cellcolor{sf_gray!18}59.82 & \cellcolor{sf_gray!18}41.47 & \cellcolor{sf_gray!18}55.02
        \\
        & SuperFlow \cite{xu2024superflow} & MinkU-101 & 96.44 & 78.57 & 56.92 & 76.29 & 54.70 & 59.35 & 71.89 & 55.13 & 60.27 & 51.60 & 60.77
        \\
        & \cellcolor{sf_yellow!8}\textbf{SuperFlow++} & \cellcolor{sf_yellow!8} MinkU-101 & \cellcolor{sf_yellow!8}\textbf{94.61} & \cellcolor{sf_yellow!8}\textbf{79.62} & \cellcolor{sf_yellow!8}\textbf{57.75} & \cellcolor{sf_yellow!8}\textbf{76.31} & \cellcolor{sf_yellow!8}\textbf{55.04} & \cellcolor{sf_yellow!8}\textbf{60.74} & \cellcolor{sf_yellow!8}\textbf{72.13} & \cellcolor{sf_yellow!8}\textbf{56.43} & \cellcolor{sf_yellow!8}\textbf{61.35} & \cellcolor{sf_yellow!8}\textbf{52.29} & \cellcolor{sf_yellow!8}\textbf{61.51}
        \\\midrule
        \multirow{5}{*}{\rotatebox[origin=c]{90}{\textbf{LP}}} & PPKT \cite{liu2021ppkt} & MinkU-34 & 183.44 & \textbf{78.15} & 30.65 & 35.42 & 28.12 & 29.21 & 32.82 & 19.52 & 28.01 & 20.71 & 28.06
        \\
        & SLidR \cite{sautier2022slidr} & MinkU-34 & 179.38 & 77.18 & 34.88 & 38.09 & 32.64 & 26.44 & 33.73 & 20.81 & 31.54 & 21.44 & 29.95
        \\
        & Seal \cite{liu2023seal} & MinkU-34 & 166.18 & 75.38 & 37.33 & 42.77 & 29.93 & 37.73 & 40.32 & 20.31 & 37.73 & 24.94 & 33.88
        \\
        & SuperFlow \cite{xu2024superflow} & MinkU-34 & 161.78 & 75.52 &  37.59 & 43.42 &  37.60 & 39.57 & 41.40 & 23.64 & 38.03 & 26.69 & 35.99
        \\
        & \cellcolor{sf_blue!8}\textbf{SuperFlow++} & \cellcolor{sf_blue!8} MinkU-34 & \cellcolor{sf_blue!8}\textbf{158.98} & \cellcolor{sf_blue!8}75.70 & \cellcolor{sf_blue!8}\textbf{38.52} & \cellcolor{sf_blue!8}\textbf{44.36} & \cellcolor{sf_blue!8}\textbf{37.91} & \cellcolor{sf_blue!8}\textbf{41.32} & \cellcolor{sf_blue!8}\textbf{41.68} & \cellcolor{sf_blue!8}\textbf{24.25} & \cellcolor{sf_blue!8}\textbf{40.22} & \cellcolor{sf_blue!8}\textbf{27.65} & \cellcolor{sf_blue!8}\textbf{36.99}
        \\\bottomrule
    \end{tabular}}
\end{table*}

\subsection{Experimental Settings}
\label{sec:settings}

\noindent\textbf{Datasets.} We conduct experiments on eleven datasets, adhering to the setup in SLidR \cite{sautier2022slidr} and Seal \cite{liu2023seal}. These datasets span a diverse range of real-world and synthetic scenarios: $^{1}$\textit{nuScenes} \cite{fong2022Panoptic-nuScenes}, $^{2}$\textit{SemanticKITTI} \cite{behley2019SemanticKITTI}, $^{3}$\textit{Waymo Open} \cite{sun2020waymoOpen}, $^{4}$\textit{ScribbleKITTI} \cite{unal2022scribbleKITTI}, $^5$\textit{RELLIS-3D} \cite{jiang2021rellis3D}, $^6$\textit{SemanticPOSS} \cite{pan2020semanticPOSS}, $^7$\textit{SemanticSTF} \cite{xiao2023semanticSTF}, $^8$\textit{SynLiDAR} \cite{xiao2022synLiDAR}, $^9$\textit{DAPS-3D} \cite{klokov2023daps3D}, $^{10}$\textit{Synth4D} \cite{saltori2022synth4D}, and $^{11}$\textit{Robo3D} \cite{kong2023robo3D}. Each dataset is evaluated using the specific data portions defined in Seal \cite{liu2023seal}. This combination ensures a robust evaluation across varied environments, label granularities, and domain characteristics.

\noindent\textbf{Implementation Details.}
The SuperFlow++ framework is implemented using the MMDetection3D \cite{mmdet3d} and OpenPCSeg \cite{pcseg2023} libraries, and all experiments are conducted on NVIDIA A100 GPUs. MinkUNet \cite{choy2019minkowski} serves as the 3D backbone, while the 2D backbones include ResNet-50 \cite{he2016resnet} and three ViT variants (ViT-S, ViT-B, and ViT-L) derived from DINOV2 \cite{oquab2023dinov2}. Semantic superpixels are generated using OpenSeeD \cite{zhang2023openSeeD}, following the methodology in Seal \cite{liu2023seal}. Pretraining is performed on 600 scenes from \textit{nuScenes} \cite{caesar2020nuScenes}, using the same splits as SLidR \cite{sautier2022slidr}. The framework is evaluated through linear probing and fine-tuning on \textit{nuScenes}, with domain generalization studies extended to the other ten datasets following Seal’s \cite{liu2023seal} configurations. Models are pretrained on 8 GPUs for 50 epochs, with a batch size of 4 per GPU and an initial learning rate of 0.01. For linear probing and fine-tuning, models are trained on 4 GPUs for 100 epochs, with a batch size of 2 per GPU; the learning rate is set to 0.01 for the task-specific head and 0.001 for the backbone. Both pretraining and fine-tuning stages adopt the AdamW optimizer \cite{loshchilov2018AdamW} with a OneCycle learning rate scheduler \cite{smith2019OneCycle} to ensure smooth convergence.

\noindent\textbf{Evaluation Protocols.}
For LiDAR semantic segmentation, we report Intersection-over-Union (IoU) per class and the mean IoU (mIoU) across all classes. For 3D object detection, performance is measured using mean Average Precision (mAP) and the nuScenes Detection Score (NDS). To assess 3D robustness, we adopt the protocols defined in Robo3D \cite{kong2023robo3D}, reporting the mean Corruption Error (mCE) and mean Resilience Rate (mRR). These metrics collectively offer a comprehensive evaluation of the framework’s effectiveness across multiple tasks and conditions.

\subsection{Comparative Study}
\label{sec:comparative_study}

\noindent\textbf{Data Pretraining Effects.} 
Linear probing serves as a standard protocol for assessing pretraining quality by freezing the pretrained 3D backbone and fine-tuning only the segmentation head. This setup isolates the representational capacity of the backbone, providing a clear indication of pretraining effectiveness. As shown in \cref{tab:benchmark}, SuperFlow++ achieves consistent improvements over state-of-the-art methods across diverse configurations. In particular, SuperFlow++ achieves an average improvement of 1.5\% mIoU over SuperFlow under the linear probing setting, demonstrating the effectiveness of our proposed designs. The performance gain primarily stems from two key components: the flow-based contrastive learning, which captures temporally coherent structural cues to enhance geometric and semantic consistency across consecutive LiDAR frames; and the temporal semantic voting strategy used during downstream evaluation, which stabilizes predictions in dynamic scenes without additional fine-tuning. Furthermore, we observe substantial gains when using larger 2D backbones. Scaling from ViT-S to ViT-L yields an increase of 1.21\% mIoU, highlighting that pretraining quality is closely tied to the representational power of the 2D network. This trend reveals a clear scaling law for image-LiDAR pretraining, suggesting that stronger 2D backbones can further enhance 3D perception performance.

\noindent\textbf{Knowledge Transfer to Downstream Tasks.}
Representation learning plays a pivotal role in reducing the dependency on large-scale annotations, making it valuable for datasets with limited labeled samples. To validate this, we compare SuperFlow++ against previous methods on three widely used datasets -- \textit{nuScenes} \cite{fong2022Panoptic-nuScenes}, \textit{SemanticKITTI} \cite{behley2019SemanticKITTI}, and \textit{Waymo Open} \cite{sun2020waymoOpen} -- under few-shot fine-tuning settings. From the results in \cref{tab:benchmark}, SuperFlow++ consistently outperforms competing approaches across all datasets, with notable improvements in low-data regimes (\textit{e.g.}, 1\% annotations), achieving approximately 1.5\% mIoU gains over the previous best method. In addition, we explore fully supervised fine-tuning on \textit{nuScenes} \cite{fong2022Panoptic-nuScenes} to assess the broader impact of pretraining. Models initialized with our pretrained backbone demonstrate significant performance advantages over randomly initialized counterparts, reaffirming the importance of pretraining for downstream 3D tasks. We also find that distillations from larger 2D backbones provide further boosts, underscoring the synergy between scalable 2D models and our pretraining framework. This result highlights the framework’s potential for future scalability and adaptability to more demanding applications.

\noindent\textbf{Domain Generalization to Other Datasets.}
The ability to generalize across domains is a critical measure of robustness for any pretraining framework. To evaluate this, we perform extensive experiments on seven diverse LiDAR datasets, each characterized by unique acquisition conditions such as adverse weather, weak annotations, synthetic data, and highly dynamic environments. As detailed in \cref{tab:multiple_datasets}, SuperFlow++ consistently surpasses prior methods, including PPKT \cite{liu2021ppkt}, SLidR \cite{sautier2022slidr}, Seal \cite{liu2023seal}, and SuperFlow \cite{xu2024superflow}, across fourteen domain generalization tasks, demonstrating its versatility and robustness. A key strength of our approach lies in its flow-based contrastive learning, which aligns features both temporally within LiDAR sequences and across LiDAR-image modalities, thereby capturing structural and semantic cues transferable across domains. Moreover, the temporal voting strategy employed during downstream evaluation enhances prediction stability under dynamic or unseen conditions. This study further underscores the adaptability of SuperFlow++ to unseen scenarios, cementing its position as a highly generalizable solution for 3D perception.

\noindent\textbf{Efficacy for 3D Object Detection.}
To further validate the efficiency of our framework, we evaluate its performance on 3D object detection tasks using the nuScenes dataset \cite{caesar2020nuScenes}. As shown in \cref{tab:detection}, SuperFlow++ consistently surpasses previous state-of-the-art methods among various data portions (5\%, 10\%, 20\%) when tested with both CenterPoint \cite{yin2021centerpoint} and SECOND \cite{yan2018second} backbones. Notably, our method achieves the highest scores under both mAP and NDS metrics, demonstrating its ability to learn effective representations even with limited labeled data. These results reaffirm the scalability and adaptability of our method to real-world 3D perception tasks, further solidifying its practical value for autonomous driving applications.

\noindent\textbf{Out-of-Distribution Robustness.}
Robustness is crucial for deploying 3D perception models in real-world autonomous driving settings, where sensing conditions often deviate from training environments. To assess this, we evaluate SuperFlow++ on the \textit{nuScenes-C} dataset from the \textit{Robo3D} benchmark \cite{kong2023robo3D}, which introduces a wide spectrum of realistic corruptions (\textit{e.g.}, adverse weather, motion blur, and sensor noise). Models are fine-tuned on the clean \textit{nuScenes} \cite{fong2022Panoptic-nuScenes} training set and then evaluated under these corrupted settings. The quantitative results in \cref{tab:robo3d} demonstrate that SuperFlow++ consistently improves robustness under diverse perturbations, confirming the effectiveness of its temporal and cross-sensor modeling strategies. Interestingly, our analysis reveals that the robustness of 3D networks also varies with their architectural capacity, with larger backbones generally demonstrating higher resilience. This finding indicates that robustness can be further improved by carefully tuning model capacity and pretraining configurations.

\begin{figure*}[t]
    \centering
    \includegraphics[width=\textwidth]{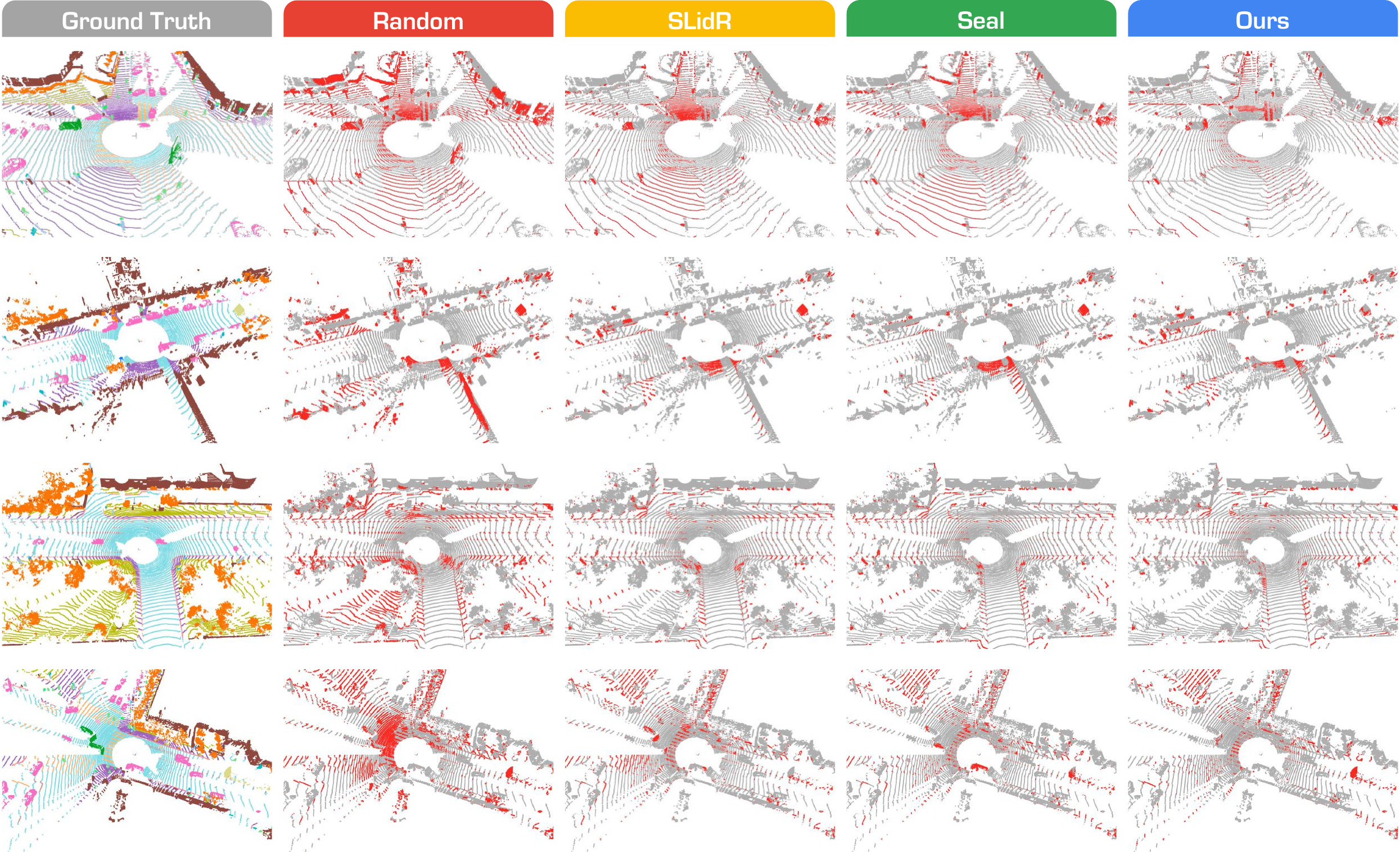}
    \vspace{-0.5cm}
    \caption{\textbf{Qualitative assessments} of state-of-the-art pretraining methods pretrained on \textit{nuScenes} \cite{caesar2020nuScenes} and fine-tuned on the \textit{Waymo Open} \cite{sun2020waymoOpen} dataset, with $1\%$ annotations. The error maps show the \textcolor{gray}{correct} and \textcolor{sf_red}{incorrect} predictions in \textcolor{gray}{gray} and \textcolor{sf_red}{red}, respectively. Best viewed in colors and zoomed-in for details.}
    \label{fig:qualitative}
\end{figure*}

\begin{figure*}[t]
    \centering
    \includegraphics[width=\textwidth]{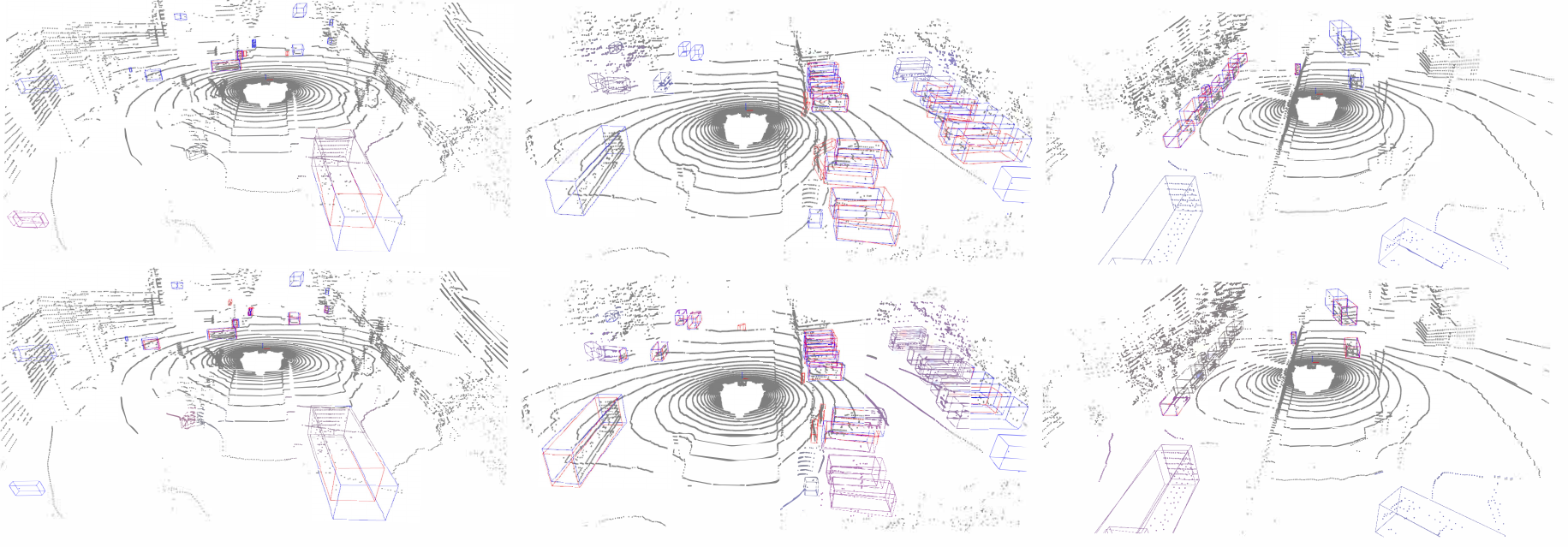}
    \vspace{-0.5cm}
    \caption{The \textbf{qualitative results} of object detection trained with 5\% labeled data. The first row shows the model trained with random initialization, while the second row displays results from our proposed framework. The \textcolor{blue}{groundtruth} / \textcolor{red}{predicted} results are highlighted with \textcolor{blue}{blue} / \textcolor{red}{red} boxes, respectively. Best viewed in colors and zoomed-in for additional details.}
    \label{fig:qualitative_detection}
\end{figure*}

\noindent\textbf{Qualitative Comparisons.}
To provide a deeper understanding of SuperFlow++’s capabilities, we present qualitative comparisons on the \textit{nuScenes}, \textit{SemanticKITTI}, and \textit{Waymo Open} datasets, benchmarked against random initialization, SLidR \cite{sautier2022slidr}, and Seal \cite{liu2023seal}. As shown in \cref{fig:qualitative}, our method consistently delivers superior segmentation performance, particularly in challenging background regions such as ``road'' and ``sidewalk''. Notably, SuperFlow++ excels in capturing intricate structural details, demonstrating robust generalization across diverse and complex scenarios. These visual comparisons highlight the model's ability to produce fine-grained, context-aware predictions, even in regions with subtle semantic distinctions. Furthermore, we evaluate SuperFlow++ on 3D object detection tasks under a few-shot setting using only 5\% annotations. As depicted in \cref{fig:qualitative_detection}, models initialized with random weights often miss objects or produce incorrect orientation predictions. In contrast, SuperFlow++ exhibits enhanced detection accuracy, reliably localizing and orienting objects even in sparse annotation scenarios. These qualitative results further validate the framework's effectiveness in addressing the challenges of 3D perception, solidifying its potential for real-world applications in autonomous driving and beyond.

\begin{figure*}[t]
    \centering
    %
    %
    \begin{subfigure}[b]{0.235\textwidth}
        \centering
        \includegraphics[width=\textwidth]{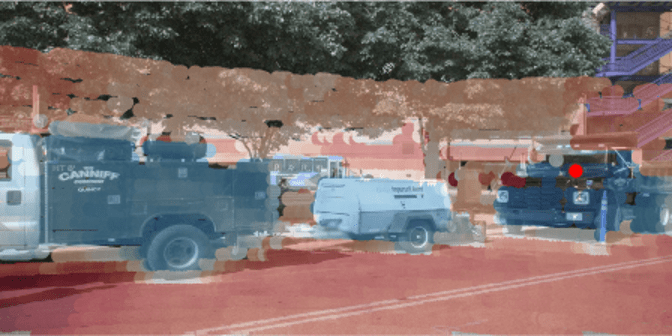}
        \caption{``car'' (3D)}
        \label{fig:car_3d}
    \end{subfigure}
    ~~
    \begin{subfigure}[b]{0.235\textwidth}
        \centering
        \includegraphics[width=\textwidth]{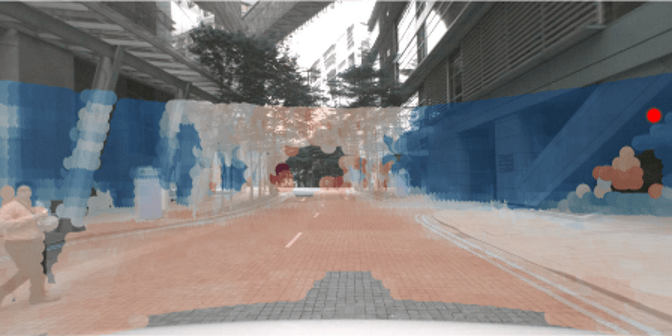}
        \caption{``manmade'' (3D)}
        \label{fig:manmade_3d}
    \end{subfigure}
    ~~
    \begin{subfigure}[b]{0.235\textwidth}
        \centering
        \includegraphics[width=\textwidth]{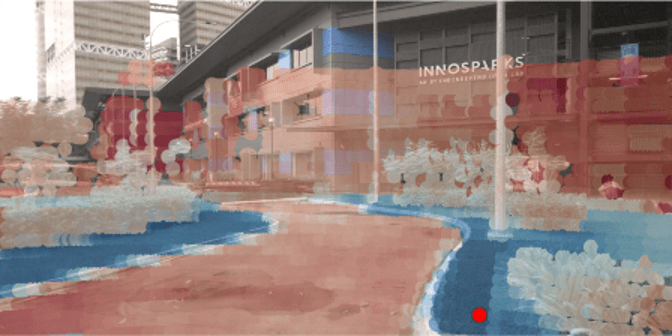}
        \caption{``sidewalk'' (3D)}
        \label{fig:sidewalk_3d}
    \end{subfigure}
    ~~
    \begin{subfigure}[b]{0.235\textwidth}
        \centering
        \includegraphics[width=\textwidth]{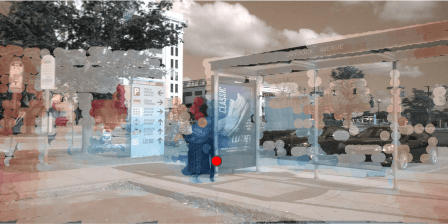}
        \caption{``flat-other'' (3D)}
        \label{fig:flatother_3d}
    \end{subfigure}
    %
    %
    \begin{subfigure}[b]{0.235\textwidth}
        \centering
        \includegraphics[width=\textwidth]{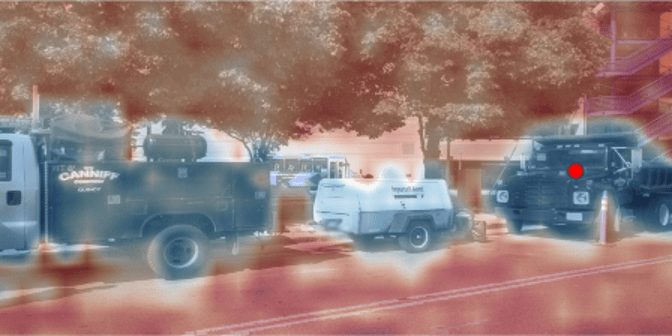}
        \caption{``car'' (2D)}
        \label{fig:car_2d}
    \end{subfigure}
    ~~
    \begin{subfigure}[b]{0.235\textwidth}
        \centering
        \includegraphics[width=\textwidth]{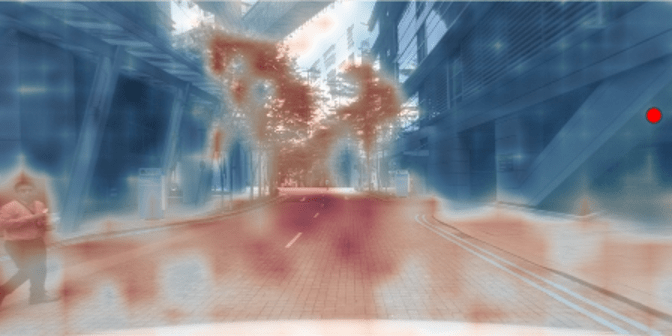}
        \caption{``manmade'' (2D)}
        \label{fig:manmade_2d}
    \end{subfigure}
    ~~
    \begin{subfigure}[b]{0.235\textwidth}
        \centering
        \includegraphics[width=\textwidth]{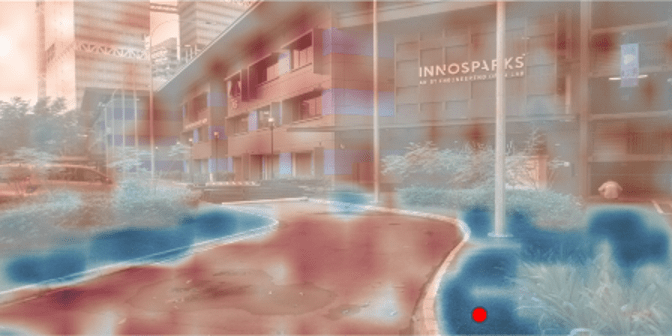}
        \caption{``sidewalk'' (2D)}
        \label{fig:sidewalk_2d}
    \end{subfigure}
    ~~
    \begin{subfigure}[b]{0.235\textwidth}
        \centering
        \includegraphics[width=\textwidth]{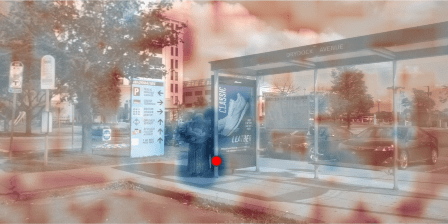}
        \caption{``flat-other'' (2D)}
        \label{fig:flatother_2d}
    \end{subfigure}
    %
    %
    \begin{subfigure}[b]{0.235\textwidth}
        \centering
        \includegraphics[width=\textwidth]{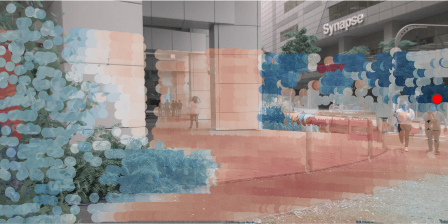}
        \caption{``vegetation'' (3D)}
        \label{fig:vegetation_3d}
    \end{subfigure}
    ~~
    \begin{subfigure}[b]{0.235\textwidth}
        \centering
        \includegraphics[width=\textwidth]{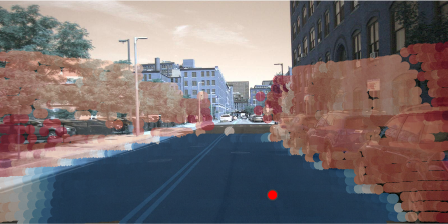}
        \caption{``driveable surface'' (3D)}
        \label{fig:driveable_3d}
    \end{subfigure}
    ~~
    \begin{subfigure}[b]{0.235\textwidth}
        \centering
        \includegraphics[width=\textwidth]{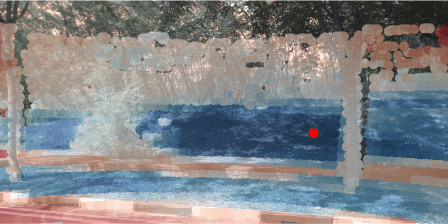}
        \caption{``terrain'' (3D)}
        \label{fig:terrain_3d}
    \end{subfigure}
    ~~
    \begin{subfigure}[b]{0.235\textwidth}
        \centering
        \includegraphics[width=\textwidth]{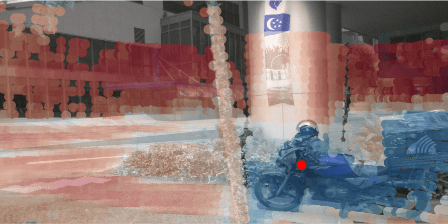}
        \caption{``motorcycle'' (3D)}
        \label{fig:motorcycle_3d}
    \end{subfigure}
    %
    %
    \begin{subfigure}[b]{0.235\textwidth}
        \centering
        \includegraphics[width=\textwidth]{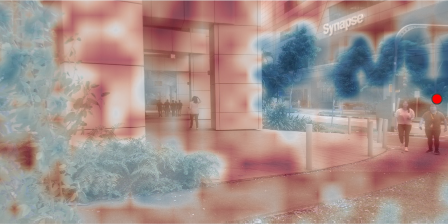}
        \caption{``vegetation'' (2D)}
        \label{fig:vegetation_2d}
    \end{subfigure}
    ~~
    \begin{subfigure}[b]{0.235\textwidth}
        \centering
        \includegraphics[width=\textwidth]{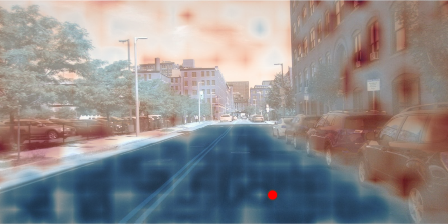}
        \caption{``driveable surface'' (2D)}
        \label{fig:driveable_2d}
    \end{subfigure}
    ~~
    \begin{subfigure}[b]{0.235\textwidth}
        \centering
        \includegraphics[width=\textwidth]{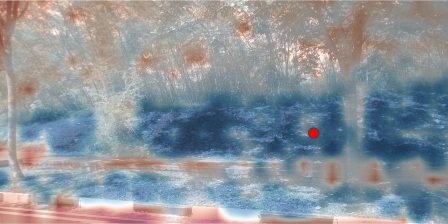}
        \caption{``terrain'' (2D)}
        \label{fig:terrain_2d}
    \end{subfigure}
    ~~
    \begin{subfigure}[b]{0.235\textwidth}
        \centering
        \includegraphics[width=\textwidth]{figures/sim_manmade2d.png}
        \caption{``motorcycle'' (2D)}
        \label{fig:motorcycle_2d}
    \end{subfigure}
    \vspace{-0.5cm}
    \caption{\textbf{Cosine similarity} between the features of a query point (denoted as a \textcolor{red}{red dot}) and the features of other points projected in the image (the 1st and 3rd rows), and the features of an image with the same scene (the 2nd and 4th rows). The color goes from \textcolor{red}{red} to \textcolor{blue}{blue} denoting low and high similarity scores, respectively.}
    \label{fig:similarity}
\end{figure*}

\subsection{Ablation Study}
\label{sec:ablation_study}

In this section, we systematically evaluate the contributions of each component in the SuperFlow++ framework. Unless specified otherwise, all experiments use MinkUNet-34 \cite{choy2019minkowski} and ViT-B \cite{oquab2023dinov2} as the 3D and 2D backbones, respectively.

\begin{table}[t]
    \centering
    \caption{\textbf{Ablation study of each component}. All variants use a MinkUNet-34 \cite{choy2019minkowski} as the 3D backbone and ViT-B \cite{oquab2023dinov2} for distillation. \textbf{VC}: View consistency. \textbf{D2S}: Dense-to-sparse regularization. \textbf{ITL}: Intra-sensor temporal contrastive learning. \textbf{CTL}: Cross-sensor temporal contrastive learning. \textbf{TV}: Temporal voting segmentation. All scores are given in percentage (\%).}
    \label{tab:ablation}
    \vspace{-0.2cm}
    \resizebox{1.0\linewidth}{!}{
    \begin{tabular}{ccccc|cc|c|c}
        \toprule
        \multirow{2}{*}{\textbf{VC}} & \multirow{2}{*}{\textbf{D2S}} & \multirow{2}{*}{\textbf{ITL}} & \multirow{2}{*}{\textbf{CTL}} & \multirow{2}{*}{\textbf{TV}} & \multicolumn{2}{c}{\textbf{nuScenes}} \vline & \textbf{KITTI} & \textbf{Waymo}
        \\
        & & & & & \textbf{LP} & \textbf{1\%} & \textbf{1\%} & \textbf{1\%}
        \\\midrule\midrule
        \multicolumn{5}{c|}{\cellcolor{sf_gray!18}Random} & \cellcolor{sf_gray!18}8.10 & \cellcolor{sf_gray!18}30.30 & \cellcolor{sf_gray!18}39.50 & \cellcolor{sf_gray!18}39.41
        \\\midrule
        \textcolor{sf_red}{\xmark} & \textcolor{sf_red}{\xmark} & \textcolor{sf_red}{\xmark} & \textcolor{sf_red}{\xmark} & \textcolor{sf_red}{\xmark} & 44.65 & 44.47 & 46.65 & 47.77
        \\
        \textcolor{sf_blue}{\cmark} & \textcolor{sf_red}{\xmark} & \textcolor{sf_red}{\xmark} & \textcolor{sf_red}{\xmark} & \textcolor{sf_red}{\xmark} & 45.57 & 45.21 & 46.87 & 48.01
        \\
        \textcolor{sf_blue}{\cmark} & \textcolor{sf_blue}{\cmark} & \textcolor{sf_red}{\xmark} & \textcolor{sf_red}{\xmark} & \textcolor{sf_red}{\xmark} & 46.17 & 46.91 & 47.26 & 49.01
        \\
        \textcolor{sf_blue}{\cmark} & \textcolor{sf_red}{\xmark} & \textcolor{sf_blue}{\cmark} & \textcolor{sf_red}{\xmark} & \textcolor{sf_red}{\xmark} & 47.24 & 47.67 & 48.21 & 49.80
        \\
        \textcolor{sf_blue}{\cmark} & \textcolor{sf_blue}{\cmark} & \textcolor{sf_blue}{\cmark} & \textcolor{sf_red}{\xmark} & \textcolor{sf_red}{\xmark} & 47.66 & 48.09 & 48.40 & 50.20
        \\
        \textcolor{sf_blue}{\cmark} & \textcolor{sf_blue}{\cmark} & \textcolor{sf_blue}{\cmark} & \textcolor{sf_blue}{\cmark} & \textcolor{sf_red}{\xmark} & 48.02 & 48.52 & 49.03 & 50.89
        \\
        \cellcolor{sf_blue!8}\textcolor{sf_blue}{\cmark} & \cellcolor{sf_blue!8}\textcolor{sf_blue}{\cmark} & \cellcolor{sf_blue!8}\textcolor{sf_blue}{\cmark} & \cellcolor{sf_blue!8}\textcolor{sf_blue}{\cmark} & \cellcolor{sf_blue!8}\textcolor{sf_blue}{\cmark} & \cellcolor{sf_blue!8}\textbf{48.86} & \cellcolor{sf_blue!8}\textbf{49.56} & \cellcolor{sf_blue!8}\textbf{49.90} & \cellcolor{sf_blue!8}\textbf{51.65}
        \\\bottomrule
    \end{tabular}}
\end{table}

\noindent\textbf{Ablative Effect of Each Component.}
We conduct an ablation study to evaluate the contributions of each component within the SuperFlow++ framework, including view consistency alignment (VC), dense-to-sparse regularization (D2S), intra-sensor temporal contrastive learning (ITL), cross-sensor temporal contrastive learning (CTL), and temporal voting segmentation (TV). The results, summarized in \cref{tab:ablation}, are based on SLiDR \cite{sautier2022slidr} with superpixels generated using a vision foundation model as the baseline. VC yields modest improvements, particularly in few-shot scenarios, by unifying semantic representations across camera views through text-prompt-guided refinement and LiDAR-to-image calibration. D2S facilitates the transfer of dense features into sparse representations and yields approximately 1.0\% mIoU gains. ITL delivers the largest performance gain (around 2.0\% mIoU) by explicitly modeling and leveraging temporal continuity across LiDAR frames. CTL further enhances temporal reasoning by aligning cross-sensor representations at adjacent timestamps, contributing an additional 0.5\% mIoU. Finally, TV aggregates semantic evidence over consecutive scans to enforce temporal prediction consistency, resulting in another 1.0\% mIoU improvement. These findings clearly illustrate the complementary contributions of each component to the overall framework, with each module addressing distinct aspects of temporal, spatial, and cross-sensor feature learning.

\begin{table}[t]
    \centering
    \caption{\textbf{Ablation study of SuperFlow} on network capacity (\# params) of 3D backbones. All methods use ViT-B \cite{oquab2023dinov2} for distillation. All scores are given in percentage (\%). Baseline results are shaded with colors. Symbol \textcolor{sf_blue}{$\bullet$} indicates the final configuration used in our model.}
    \vspace{-0.2cm}
    \label{tab:network_capacity}
    \resizebox{1.0\linewidth}{!}{
    \begin{tabular}{r|c|cc|c|c|cc}
        \toprule
        \multirow{2}{*}{\textbf{Backbone}} & \multirow{2}{*}{\textbf{L}} & \multicolumn{2}{c|}{\textbf{nuScenes}} & \textbf{KITTI} & \textbf{Waymo} & \multicolumn{2}{c}{\textbf{nuScenes-C}}
        \\
        & & \textbf{LP} & \textbf{1\%} & \textbf{1\%} & \textbf{1\%} & \textbf{mCE} & \textbf{mRR}
        \\\midrule\midrule
        MinkUNet~\textcolor{sf_blue}{$\circ$} & 18 & 47.20 & 47.70 & 48.04 & 49.24 & 109.00 & 75.66
        \\
        \cellcolor{sf_blue!8}MinkUNet~\textcolor{sf_blue}{$\bullet$} & \cellcolor{sf_blue!8}34 & \cellcolor{sf_blue!8}47.66 & \cellcolor{sf_blue!8}48.09 & \cellcolor{sf_blue!8}48.40 & \cellcolor{sf_blue!8}50.20 & \cellcolor{sf_blue!8}91.67 & \cellcolor{sf_blue!8}83.17
        \\
        MinkUNet~\textcolor{sf_blue}{$\circ$} & 50 & 54.11 & 52.86 & 49.22 & 51.20 & 107.35 & 74.02
        \\
        MinkUNet~\textcolor{sf_blue}{$\circ$} & 101 & 52.56 & 51.19 & 48.51 & 50.01 & 96.44 & 78.57
        \\\bottomrule
    \end{tabular}}
\end{table}

\noindent\textbf{3D Network Capacity.}
The capacity of 3D networks is a pivotal factor in effective representation learning, especially given their smaller scale compared to 2D backbones. To investigate this, we evaluate the performance of different MinkUNet architectures, with results detailed in \cref{tab:network_capacity}. The findings reveal a consistent performance improvement as the network size increases, except for MinkUNet-101, where overparameterization appears to impede convergence. This suggests that smaller networks lack the capacity to capture complex patterns effectively, while excessively large networks encounter optimization challenges during pretraining. These results underscore the importance of carefully balancing model capacity to achieve optimal performance in 3D representation learning. 

\begin{table}[t]
    \centering
    \caption{\textbf{Ablation study of SuperFlow} using different \# of sweeps. All methods use ViT-B \cite{oquab2023dinov2} for distillation. All scores are given in percentage (\%). Baseline results are shaded with colors. The memory is measured with a batch size of 1. Symbol \textcolor{sf_blue}{$\bullet$} indicates the final configuration used in our model.}
    \vspace{-0.2cm}
    \label{tab:sweeps}
    \scalebox{1.0}{
    \begin{tabular}{r|c|cc|c|c}
        \toprule
        \multirow{2}{*}{\textbf{Sweeps}} & \multirow{2}{*}{\textbf{Memory}} & \multicolumn{2}{c}{\textbf{nuScenes}} \vline & \textbf{KITTI} & \textbf{Waymo}
        \\
        & & \textbf{LP} & \textbf{1\%} & \textbf{1\%} & \textbf{1\%}
        \\\midrule\midrule
        $1\times$ Sweeps~\textcolor{sf_blue}{$\circ$} & 5.7 GB & 47.41 & 47.52 & 48.14 & 49.31
        \\
        \cellcolor{sf_blue!8}$2\times$ Sweeps~\textcolor{sf_blue}{$\bullet$} & \cellcolor{sf_blue!8}5.9 GB & \cellcolor{sf_blue!8}47.66 & \cellcolor{sf_blue!8}48.09 & \cellcolor{sf_blue!8}48.40 & \cellcolor{sf_blue!8}50.20
        \\
        $5\times$ Sweeps~\textcolor{sf_blue}{$\circ$} & 6.3 GB & 47.23 & 48.00 & 47.94 & 49.14
        \\
        $7\times$ Sweeps~\textcolor{sf_blue}{$\circ$} & 7.0 GB & 46.03 & 47.98 & 46.83 & 47.97
        \\\bottomrule
    \end{tabular}}
\end{table}

\noindent\textbf{Representation Density.}
Dense-to-sparse regularization is central to SuperFlow++, ensuring consistency across point clouds of varying densities. To investigate its impact, we evaluate the framework under different density configurations, as summarized in \cref{tab:sweeps}. The results reveal that moderate densities -- achieved by merging two or three LiDAR sweeps -- offer optimal performance. This configuration effectively balances spatial richness and temporal coherence, enable the model to capture detailed yet consistent features across scenes. In contrast, excessive density, resulting from aggregating too many sweeps, introduces significant challenges. Motion artifacts arising from dynamic objects create spatial distortions, leading to misalignments between sparse superpixels and their corresponding dense superpoints. Such misalignments degrade the quality of learned representations, as the framework struggles to reconcile conflicting spatial and temporal cues. These findings highlight the critical need for selecting an optional density level to maximum the benefits of the regularization with avoiding potential drawbacks.

\begin{table}[t]
    \centering
    \caption{\textbf{Ablation study on spatiotemporal consistency}. All variants employ MinkUNet-34 \cite{choy2019minkowski} as the 3D backbone and ViT-B \cite{oquab2023dinov2} as the distillation teacher. The notation $\mathbf{0}$ represents the current timestamp, while $\mathbf{0.5s}$ corresponds to a temporal offset of $0.5$ seconds at a sampling frequency of $20$ Hz. All scores are given in percentage (\%).}
    \vspace{-0.2cm}
    \label{tab:temporal}
    \resizebox{\linewidth}{!}{
    \begin{tabular}{c|cccc|c|c}
        \toprule
        \multirow{2}{*}{\textbf{Timespan}} & \multicolumn{4}{c}{\textbf{nuScenes}} \vline & \textbf{KITTI} & \textbf{Waymo}
        \\
        & \textbf{LP} & \textbf{1\%} & \textbf{5\%} & \textbf{10\%} & \textbf{1\%} & \textbf{1\%}
        \\\midrule\midrule
        \cellcolor{sf_gray!18}Single-Frame & \cellcolor{sf_gray!18}46.17 & \cellcolor{sf_gray!18}46.91 & \cellcolor{sf_gray!18}57.86 & \cellcolor{sf_gray!18}62.75 & \cellcolor{sf_gray!18}47.26 & \cellcolor{sf_gray!18}49.01
        \\\midrule
        0, -0.5s & 46.39 & 47.08 & 58.35 & 63.81 & 47.99 & 49.78
        \\
        \cellcolor{sf_blue!8}-0.5s, 0, +0.5s & \cellcolor{sf_blue!8}47.66 & \cellcolor{sf_blue!8}48.09 & \cellcolor{sf_blue!8}59.66 & \cellcolor{sf_blue!8}64.52 & \cellcolor{sf_blue!8}48.40 & \cellcolor{sf_blue!8}50.20
        \\
        -1.0s, 0, +1.0s & 47.60 & 47.99 & 59.24 & 64.32 & 48.43 & 50.18
        \\
        -1.5s, 0, +1.5s & 46.43 & 48.27 & 59.18 & 64.35 & 48.34 & 49.93
        \\
        -2.0s, 0, +2.0s & 46.20 & 48.49 & 59.03 & 64.12 & 48.18 & 50.01
        \\\bottomrule
    \end{tabular}}
\end{table}

\noindent\textbf{Temporal Consistency.}
Leveraging temporal cues is essential for fully exploiting the sequential nature of LiDAR data. Temporal contrastive learning, as formulated in \cref{eq:temporal}, enforces semantic coherence by aligning superpoint features across consecutive scenes. Results presented in \cref{tab:temporal} demonstrate that this approach significantly outperforms single-frame methods, underscoring the critical role of temporal information in enhancing feature learning. We further investigate the influence of two factors: the number of frames and the timespan between them. Using three frames yields superior performance compared of two frames, as it captures richer context-aware information and provides a broader temporal perspective. However, an increase in the timespan between frames leads to a noticeable decline in performance. This suggests that shorter timespans preserve scene continuity and maintain consistent semantic relationships, while longer timespans introduce greater variability in object positions and scene dynamics, making temporal alignment more challenging and less effective.

\begin{table}[t]
    \centering
    \caption{\textbf{Effect of SuperFlow++ on semi-supervised learning}. Performance is evaluated using various portions of labeled data, while the remaining data is treated as unlabeled. The semi-supervised learning setup follows the LaserMix~\cite{kong2023lasermix} paradigm. All scores are given in percentage (\%).}
    \label{tab:semi}
    \vspace{-0.2cm}
    \resizebox{1.0\linewidth}{!}{
    \begin{tabular}{r|cccc|c|c}
        \toprule
        \multirow{2}{*}{\textbf{Methods}} & \multicolumn{4}{c|}{\textbf{nuScenes}} & \textbf{KITTI} & \textbf{Waymo}
        \\
        & \textbf{1\%} & \textbf{5\%} & \textbf{10\%} & \textbf{25\%} & \textbf{1\%} & \textbf{1\%}
        \\\midrule\midrule
        Random & 30.30 & 47.84 & 56.15 & 65.48 & 39.50 & 39.41
        \\
        SuperFlow++ & 49.56 & 60.75 & 65.46 & 70.19 & 49.90 & 51.65
        \\
        \cellcolor{sf_blue!8}\textit{w/} Semi-sup. & \cellcolor{sf_blue!8}\textbf{53.55} & \cellcolor{sf_blue!8}\textbf{65.49} & \cellcolor{sf_blue!8}\textbf{68.25} & \cellcolor{sf_blue!8}\textbf{72.48} & \cellcolor{sf_blue!8}\textbf{54.42} & \cellcolor{sf_blue!8}\textbf{55.42}
        \\\bottomrule
    \end{tabular}}
\end{table}

\noindent\textbf{Semi-Supervised Downstream Performance.}
To fully leverage the abundance of LiDAR data, we extend SuperFlow++ to the semi-supervised setting, where only a fraction of the data is labeled while the remaining samples are unlabeled, following the paradigm of LaserMix \cite{kong2023lasermix}. As shown in \cref{tab:semi}, our approach achieves a significant improvement across different label proportions, demonstrating its effectiveness in low-data regimes. Notably, even with a limited fraction of labeled data, SuperFlow++ consistently outperforms baseline methods, highlighting its ability to extract meaningful representations from unlabeled LiDAR scans. This suggests that the integration of spatiotemporal constraints and cross-sensor distillation enhances feature learning in a label-efficient manner. Furthermore, as the proportion of labeled data increases, our method continues to scale effectively, closing the gap to fully supervised performance.

\noindent\textbf{Visual Inspections.}
To further assess the qualitative impact of SuperFlow++, we visualize similarity maps of pretrained features for different semantic categories. As shown in \cref{fig:similarity}, SuperFlow++ produces highly coherent similarity responses within regions sharing the same semantic label as the reference point, indicating strong intra-class compactness and inter-class separability in the learned embedding space, which directly benefits downstream tasks. This performance stems from three key core design elements: 1) view-consistent superpixels enable the model to learn unified semantic representations by reducing modality-specific biases, 2) dense-to-sparse regularization ensures robust generalization by allowing the model to adapt seamlessly to varying point cloud densities, which are common in real-world scenarios, and 3) temporal contrastive learning captures semantic and contextual relationships across sequential scenes, enhancing the quality of learned representations.

\section{Conclusion}
\label{sec:conclusion}

In this work, we present \textbf{SuperFlow++}, a novel framework for spatiotemporal-consistent 3D representation learning. By leveraging the sequential nature of LiDAR data, we introduce four key components: view consistency alignment, dense-to-sparse regularization, flow-based contrastive learning, and temporal voting for semantic segmentation. These strategies enhance the temporal stability and semantic fidelity of learned representations. Through comprehensive experiments across multiple LiDAR datasets, SuperFlow++ consistently outperforms existing methods in various evaluation settings, including linear probing, fine-tuning, and robustness assessments. Additionally, our study on scaling 2D and 3D network capabilities provides new insights into the trade-offs between model complexity and representation quality, offering guidance for future advancements in large-scale 3D pretraining. Beyond improving 3D perception accuracy, SuperFlow++ serves as a foundation for next-generation 3D learning frameworks, with the potential to bridge gaps between spatial and temporal understanding in autonomous systems. We anticipate that future research will explore and refine these concepts, driving progress toward more efficient and generalizable 3D vision models.


\ifCLASSOPTIONcaptionsoff
  \newpage
\fi
\bibliographystyle{IEEEtran}
\bibliography{ref}

\end{document}